%% file: paper.tex
\documentclass[10pt,logo,copyright]{nvidiatechreport}
\input{packages}

\usepackage[symbol]{footmisc}

\renewcommand{\today}{2026-06-08}

\usepackage[group-separator = {,},  
            group-minimum-digits = 4,  
            group-digits = integer     
           ]{siunitx}

\definecolor{darkred}{rgb}{0.7, 0.0, 0.0}

\definecolor{golden}{RGB}{229, 184, 11}

\usepackage{tikz}
\usetikzlibrary{arrows.meta,calc}

\usepackage{pifont}

\usepackage[nameinlink]{cleveref}
\crefname{equation}{Eq.}{Eqs.}
\crefname{figure}{Fig.}{Figs.}
\crefname{section}{Sec.}{Sec.}
\crefname{appendix}{App.}{App.}
\crefname{table}{Tab.}{Tabs.}
\crefname{algorithm}{Algo}{Algo}
\crefname{thm}{Thm}{Thm}
\Crefname{thm}{Thm}{Thm}
\crefname{prop}{Prop}{Prop}
\usepackage{tcolorbox}
\usepackage{wrapfig}
\usepackage{twemojis}
\tcbuselibrary{breakable} 
\tcbset{
  colback=gray!5!white,
  colframe=gray!60!black,
  boxrule=0.6pt,
  arc=3pt
}

\newtcolorbox{promptbox}[1][]{%
  title=#1,
  coltitle=black,
  fonttitle=\bfseries,
  colbacktitle=gray!15,
  breakable,
}

\usepackage{listings}
\usepackage{courier}
\usepackage{lettrine}

\usepackage{microtype}
\usepackage{tcolorbox}
\usepackage{adjustbox}
\usepackage{threeparttable}
\usepackage{graphicx}
\definecolor{gold}{HTML}{FFD700}

\definecolor{metallicgold}{HTML}{D4AF37}

\newcommand{\crefnames}[3]{%
  \@for\next:=#1\do{%
    \expandafter\crefname\expandafter{\next}{#2}{#3}%
  }%
}

\newcommand{\hflogo}{\raisebox{-0.2\height}{\includegraphics[height=1em]{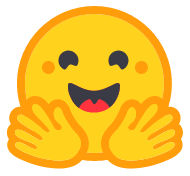}}}

\title{Unified Audio Intelligence Without Regressing on Text Intelligence}

\definecolor{c0}{cmyk}{1,0.3968,0,0.2588} 
\definecolor{c1}{cmyk}{0,0.6175,0.8848,0.1490} 
\definecolor{c2}{cmyk}{0.1127,0.6690,0,0.4431} 
\definecolor{c3}{cmyk}{0.3081,0,0.7209,0.3255} 
\colorlet{c3light}{c3!30!white}
\definecolor{nvg}{HTML}{33CC00}
\newtcbox{\hlprimary}{on line,colback=c0!10,colframe=white,size=fbox,arc=3pt, box align=base,before upper=\strut, top=-2pt, bottom=-4pt, left=-1pt, right=-1pt, boxrule=0pt}
\newtcbox{\hlprimarytab}{on line, box align=base, colback=c0!10,colframe=white,size=fbox,arc=3pt, before upper=\strut, top=-2pt, bottom=-4pt, left=-2pt, right=-2pt, boxrule=0pt}
\newtcbox{\hlsecondary}{on line,colback=c1!10,colframe=white,size=fbox,arc=3pt, box align=base,before upper=\strut, top=-2pt, bottom=-4pt, left=-1pt, right=-1pt, boxrule=0pt}
\newtcbox{\hlsecondarytab}{on line, box align=base, colback=c1!10,colframe=white,size=fbox,arc=3pt, before upper=\strut, top=-2pt, bottom=-4pt, left=-2pt, right=-2pt, boxrule=0pt}
\newtcolorbox{hlmultiline}{on line,colback=decentgrey!75,colframe=white,size=fbox,arc=3pt, box align=base, top=0pt, bottom=2pt, boxrule=0pt, before=\adjustbox{valign=c}\bgroup, after=\egroup, before upper=\strut}
\newtcbox{\hlmaintab}{on line, box align=base, colback=nvg!25,colframe=white,size=fbox,arc=3pt, before upper=\strut, top=-2pt, bottom=-4pt, left=-2pt, right=-2pt, boxrule=0pt}

\newcolumntype{Y}{>{\centering\arraybackslash}X}
\newcolumntype{Z}{>{\raggedleft\arraybackslash}X}

\definecolor{c4}{cmyk}{0.6765,0.2017,0,0.0667} 
\definecolor{c5}{cmyk}{0,0.8765,0.7099,0.3647} 

\definecolor{darkgrey}{RGB}{149,149,149}
\definecolor{decentgrey}{RGB}{242,242,242}

\newcommand{\longname}{Nemotron-Labs-Audex\xspace}
\newcommand{\model}{Audex\xspace}

\author{
Zhifeng Kong\footnote[1]{Equal contribution, with authors listed in reverse alphabetical order by first name.},  
~Sang-gil Lee$^*$,  ~Jaehyeon Kim$^*$,  ~Boxin Wang$^*$,  ~Zihan Liu, ~Sungwon Kim, ~Yang Chen, \\
\small\textbf{Arushi Goel, ~Rajarshi Roy, ~Wenliang Dai, ~Zhuolin Yang, ~Yangyi Chen, ~Dongfu Jiang, ~Sreyan Ghosh,} \\
\vspace{.05cm}
\small\textbf{Tuomas Rintamaki, ~Andrew Tao, ~Jonathan Raiman, ~Mohammad Shoeybi, ~Bryan Catanzaro, ~Wei Ping$^*$}\footnote[2]{Leads the effort. Correspondence to: \{zkong, sanggill, jaehyeonk, boxinw,  wping\}@nvidia.com.}
}

\begin{abstract}
{\normalfont
Audio intelligence involves understanding, reasoning about, and generating both audio and speech.
In this work, we introduce Nemotron-Labs-Audex-30B-A3B (Audex), a unified audio-text LLM built on Nemotron-Cascade-2-30B-A3B, a strong text-only MoE LLM.
Audex adopts a simple unified design with a single Transformer decoder: audio inputs are encoded and projected into the text embedding space, while text tokens and quantized audio output tokens are treated uniformly during generation.
This architecture enables strong audio-text fusion, seamless multimodal generation, and compatibility with standard LLM training and inference infrastructure.
For training, we meticulously curate audio-text datasets comprising 157.4B audio tokens and 320.5B text tokens. We apply multi-stage supervised training on these datasets, followed by text-only Cascade RL and multi-domain on-policy distillation.
\model delivers state-of-the-art audio understanding, speech recognition and translation, text-to-speech, audio generation, and speech-to-speech generation, while preserving very compelling reasoning,  alignment, knowledge, long-context, and agentic capabilities of its text-only LLM backbone with marginal or no regression.
We release the model checkpoints to facilitate open research.
}
\end{abstract}

\begin{document}

\maketitle

\abscontent

\begin{itemize}
    \item[\hflogo] \href{https://huggingface.co/nvidia/Nemotron-Labs-Audex-30B-A3B}{\longname-30B-A3B}: the audio-text to audio-text model.
    \item[\hflogo] \href{https://huggingface.co/nvidia/Nemotron-Labs-Audex-2B}{\longname-2B}: the smaller 2B model uses the same training recipe as the 30B-A3B model.
\end{itemize}

\vspace{.2cm}



\renewcommand{\thefootnote}{\arabic{footnote}}

\input{sections/1_Introduction}

\input{sections/2_Main_results}

\input{sections/4_Model_Architecture}

\input{sections/5_Training_Recipes}

\input{sections/6_Results}

\input{sections/3_Related_Work}

\section{Acknowledgments}

We would like to extend our gratitude to the broader NVIDIA team for valuable discussions. 
We thank Lukas Voegtle and Philipp Fischer for helpful discussions on the Megatron-Energon dataloader.
We thank Boris Ginsburg, Zhehuai Chen, Jason Li, Edresson Casanova, and Yongqiang Wang for helpful discussions on building speech LLMs. 
We also thank Jinchuan Tian for his exploratory work during his internship.

\clearpage

\paragraph{\LARGE Appendix}

\appendix

\input{sections/Appendix_Benchmarks}

\input{sections/Appendix}

\clearpage
\bibliographystyle{plainnat}
\bibliography{paper}

\end{document}

%% file: packages.tex
\usepackage[authoryear,round,sort]{natbib}
\usepackage[utf8]{inputenc} 
\usepackage[T1]{fontenc}    

\usepackage{parskip}        
\usepackage{url}            
\usepackage{booktabs}       
\usepackage{amsfonts}       
\usepackage{nicefrac}       
\usepackage{microtype}      
\usepackage{xcolor}         
\usepackage[dvipsnames]{xcolor} 
\usepackage[table]{xcolor}
\usepackage{graphicx}
\usepackage{animate}        
\usepackage{subcaption}
\usepackage{tabularx}
\usepackage{makecell}
\usepackage{adjustbox}
\usepackage{array}
\usepackage{setspace}
\newcolumntype{M}[1]{>{\centering\arraybackslash}m{#1}}
\usepackage{float}
\usepackage{tikz}
\usetikzlibrary{positioning,shapes,arrows}
\usepackage{amsmath,amsfonts,bm, bbm,leftindex}
\usepackage{multirow}
\usepackage{comment}
\usepackage{gensymb}
\usepackage{lipsum}
\usetikzlibrary{arrows.meta, positioning, fit}
\usepackage[para]{threeparttable}
\usepackage{tikz}
\usepackage{fvextra}
\usepackage[most]{tcolorbox}
\usepackage{adjustbox} 
\usepackage{pgfplots}

\usepackage{marginnote} 
\pgfplotsset{compat=1.18}
\usepackage{subcaption}
\usetikzlibrary{tikzmark}

\newenvironment{TableSmall}
{%
  \footnotesize
  \setlength{\tabcolsep}{4.5pt}
  \renewcommand{\arraystretch}{0.95}
}
{}


%% file: sections/1_Introduction.tex
\section{Introduction}
\label{sec:introduction}

Audio, including speech, music and environmental sound, is central to how humans perceive, communicate, and experience the world, from understanding the physical environment and engaging in spoken conversation to expressing emotions beyond written language and appreciating music.
As such, audio intelligence, which involves understanding, reasoning about, and generating both audio and speech, is an indispensable modality for building artificial general intelligence~(AGI).

In recent years, substantial efforts have been devoted to building audio LLMs that excel at audio and speech understanding, including Audio Flamingo~\citep{kong2024audio, ghosh2026audio,ghosh2025audio,goel2025audio}, Qwen2-Audio~\citep{chu2024qwen2audio}, Kimi-Audio~\citep{ding2025kimi}, Voxtral \citep{liu2025voxtral}, and Step-Audio 2~\citep{wu2025step}. Other works have explored unified LLMs capable of both audio understanding and generation, including Qwen-Omni~\citep{xu2025qwen2,xu2025qwen3omni,qwen3.5_omni}, MiMo-Audio \citep{zhang2025mimo}, and UALM~\citep{tian2025ualm}.

\begin{samepage}
It is worth noting that LLMs capable of handling multimodal inputs and outputs, including audio and vision, especially those supporting multimodal generation, often exhibit noticeable regressions on text benchmarks.
For vision-only inputs, the NVLM study~\citep{dai2024nvlm} documented this issue and addressed it by training on a mixture of high-quality text-only and vision-text data.
For models with multimodal output capabilities, the issue is even more challenging. 
For example, even when the multimodal output domain is constraint to speech generation, Qwen3-Omni~\citep{xu2025qwen3omni} and Qwen3.5-Omni~\citep{qwen3.5_omni} show degradation on key text benchmarks relative to their text-only counterparts -- Qwen3~\citep{yang2025qwen3} and Qwen3.5~\citep{qwen35blog}.~\footnote{For example, Table~5 of the Qwen3-Omni technical report compares Qwen3-30B-A3B-Thinking-0527 with Qwen3-Omni-30B-A3B-Thinking and shows noticeable regressions on reasoning benchmarks.} 
Such degradation is concerning because much of a model’s intelligence, including reasoning ability, knowledge, and agentic capabilities such as tool use, is encoded in text.
\end{samepage}

The challenge is well recognized in frontier LLM development: building native multimodal LLMs capable of multimodal understanding and generation without compromising text intelligence is arguably a critical step on the path toward AGI. 
Along this direction, several efforts have integrated audio and vision understanding and generation into frontier proprietary LLMs, with GPT-4o~\citep{hurst2024gpt} and Gemini 2.5~\citep{comanici2025gemini} as notable examples. 
In contrast, open frontier LLMs, such as DeepSeek-V4~\citep{deepseekai2026deepseekv4} and GLM-5.1~\citep{glm5team2026glm5vibecodingagentic}, remain text-only. 
Although Qwen3.5~\citep{qwen35blog} and Kimi-2.5~\citep{team2026kimi} incorporate vision inputs, they do not yet support multimodal outputs such as audio.

In this work, we build and release \longname-30B-A3B~(\textbf{\model} for short throughout the paper) on top of the text-only Nemotron-Cascade-2-30B-A3B~\citep{Nemotron_Cascade_2}, a strong general-purpose LLM based on a 30B Mamba2-Transformer hybrid Mixture-of-Experts (MoE) architecture with 3B activated parameters~\citep{nemotron_nano_v3}.
\model achieves state-of-the-art results in general audio understanding and generation across speech, music, and sound, while preserving the text intelligence of its LLM backbone, Nemotron-Cascade 2, with marginal or no regression~(see Table~\ref{tab:main_results}).
To this end, Audex is trained on the same text-only post-training data used for Nemotron-Cascade 2, combined with audio-text data.

\model is a single unified model that supports both \emph{instruct} mode for instant responses and \emph{thinking} mode for reasoning across both text-only and audio-to-text tasks.
Unlike cascaded systems with stacked models~\citep{thinkingmachines2026interactionmodels} or thinker–talker architectures~\citep{xu2025qwen3omni}, \model is a single MoE Transformer decoder model that ingests audio inputs through an audio encoder followed by an MLP projection into the same embedding space as text, while treating text tokens and quantized audio tokens uniformly during autoregressive generation.
This unified audio-text architecture enables seamless \emph{modality fusion} and maximizes capabilities in audio-related tasks while preserving text intelligence across a broad range of benchmarks, especially reasoning benchmarks.
Furthermore, the simple architecture is readily compatible with existing LLM training infrastructure~(e.g., Megatron-LM~\citep{shoeybi2019megatron}) and inference stacks~(e.g., vLLM~\citep{kwon2023efficient}), enabling scalable training and inference through well-engineered implementations.

We organize the remainder of this technical report as follows. 
We first highlight the main results in Section~\S\ref{sec:main_results}. 
We then introduce the model architecture and components in Section~\S\ref{sec:model_arch}, followed by the model training recipes in Section~\S\ref{sec:training}. 
In Section~\S\ref{sec:detailed_results}, we present detailed results across different domains and training stages, along with more in-depth analysis.
Finally, we discuss related work in Section~\S\ref{sec:related_work}.

%% file: sections/2_Main_results.tex
\section{Main Results}
\label{sec:main_results}

\begin{table}[t!]
\centering
\footnotesize
\renewcommand{\arraystretch}{1.2}
\caption{\textbf{Main results}. 
We report the results of Nemotron-Lab-Audex-30B-A3B alongside those of its text-only backbone, Nemotron-Cascade-2-30B-A3B.
For the baseline models and their text-only backbones, we use numbers from the official report when available; otherwise, we evaluate them, as indicated by $^\ddag$, using the recommended settings.
We mark tool-integrated reasoning results with $\diamondsuit$ for math and code reasoning.
}
\label{tab:main_results}
\begin{adjustbox}{width={1.0\textwidth}}
\begin{tabular}{l|c|cc|cc|c>{\columncolor{c3light}}c}
\toprule
\shortstack[l]{\textbf{Benchmark} \\ \textbf{} \\ \textbf{}}
 & \shortstack{\textbf{Step-Audio}\\ \textbf{R1.1} \\ \textbf{33B}}
  & \shortstack{\textbf{Qwen3}\\ \textbf{30B-A3B} \\ \textbf{Thinking-2507}}
 & \shortstack{\textbf{Qwen3-Omni}\\ \textbf{30B-A3B} \\ \textbf{Thinking}}
  & \shortstack{\textbf{Qwen3.5}\\ \textbf{35B-A3B} \\ \textbf{}}
 & \shortstack{\textbf{Qwen3.5}\\ \textbf{Omni-Flash} \\ {(Proprietary)}}
  & \shortstack{\textbf{Nemotron}\\ \textbf{Cascade-2} \\ \textbf{30B-A3B}}
 & \shortstack{\textbf{{\model}}\\ \textbf{{30B-A3B}} \\ \textbf{} } 
 \\
\midrule
Context Length                      & 64K & 256K & 64K & 256K & 256K & 1M & 1M \\ 
\midrule
\multicolumn{5}{l}{\textbf{Reasoning}} \\
AIME 2025 & 44.8$^\ddag$ & 85.0 & 73.7 & 91.9$^\ddag$ & -- & 92.4|98.6$^\diamondsuit$ & 91.2|98.3$^\diamondsuit$ \\
AIME 2026 & 57.8$^\ddag$ & -- & -- & 91.1$^\ddag$  & -- & 90.9|95.0$^\diamondsuit$ & 89.4|96.6$^\diamondsuit$ \\
HMMT Feb25 & 27.0$^\ddag$ & 71.4 & 60.4$^\ddag$ & 89.0 & 59.0 & 94.6|92.2$^\diamondsuit$  &  92.2|93.8$^\diamondsuit$  \\
IMO AnswerBench & -- &  65.7$^\ddag$ & 59.9$^\ddag$ & 74.8$^\ddag$ & 51.5 & 79.3 & 81.1 \\
LiveCodeBench v6  & 22.9$^\ddag$  &  66.0  & 59.2$^\ddag$ & 74.6  & 56.6 & 87.2|88.4$^\diamondsuit$  &  85.3|86.2$^\diamondsuit$  \\
SciCode            & --  & 33.3 & 30.6 & 38.0  & 25.5 &  36.4 & 35.2 \\
\midrule
\multicolumn{5}{l}{\textbf{Knowledge}} \\
MMLU-Redux         & 86.6$^\ddag$  &  91.4 & 88.8 & 93.3 & 90.0 & 86.3 & 86.4 \\
MMLU-Pro           & 75.3$^\ddag$ & 80.9 & 80.4$^\ddag$ & 85.3 & 79.9 &  79.8 &  78.9 \\
GPQA-Diamond       & 60.7$^\ddag$ & 73.4 & 73.1 & 84.2  & 76.4 & 76.1 &  74.9 \\
\midrule
\multicolumn{5}{l}{\textbf{Alignment}} \\
ArenaHard v2      & 44.3$^\ddag$ & 56.0  & 55.1   &  65.4$^\ddag$ & -- & 83.5 & 81.6 \\
IFBench (prompt)  & 32.2$^\ddag$ & 62.2$^\ddag$ & 52.4$^\ddag$ & 70.2 & 38.4 & 82.9 & 77.8 \\
\midrule
\multicolumn{5}{l}{\textbf{Long Context }} \\
AA-LCR                  & N/S & 51.8$^\ddag$ & 11.0$^\ddag$ & 58.5  & 46.0 & 39.1 & 39.3 \\
LongBench v2            & N/S  & 47.2$^\ddag$ & 27.2$^\ddag$ & 59.0 & 46.4 & 40.3 & 41.3 \\
NIAH (256K|1M) & N/S & 95.2|19.0$^\ddag$ & 0.0|0.0$^\ddag$ & -- & -- & 100|99.0 & 99.4|83.4 \\
\midrule
\multicolumn{5}{l}{\textbf{Agentic}} \\
$\tau^2$-Bench              & N/S &  47.7  & 45.4 & 81.2  & 78.0 & 58.9 & 57.2 \\
Terminal Bench 2.0          & N/S & 2.2$^\ddag$ & 1.1$^\ddag$ & 40.5  & -- & 21.1  & 19.1 \\
SWE Verified       & N/S & 22.8$^\ddag$ & 23.4$^\ddag$ & 69.2  & -- & 50.2 & 48.2 \\ 
\midrule
\midrule
\multicolumn{5}{l}{\textbf{Audio Understanding}} \\
MMAU                    & 73.6  & N/S  & 75.4 & N/S  & 80.4 & N/S & 75.6 \\ 
MMAR                    & 69.8  & N/S & 66.4 &  N/S & 74.0 & N/S & 63.2 \\ 
MMSU                    & 74.1  & N/S & 70.2 & N/S  & 72.2 & N/S & 63.4 \\ 
Audio Entailment$_2$ & 61.6 & N/S & 61.6 & N/S & -- & N/S & 95.0 \\
\midrule
\multicolumn{5}{l}{\textbf{Speech Recognition (WER$\downarrow$)}} \\ 
OpenASR                    & 7.91$^\ddag$ & N/S & 8.00$^\ddag$ &  N/S & -- & N/S & 6.82 \\ 
LibriSpeech (clean)  & 1.66$^\ddag$  & N/S  &  2.22$^\ddag$ & N/S & 1.11 & N/S & 1.34 \\ 
LibriSpeech (other)  & 3.28$^\ddag$  & N/S  & 4.38$^\ddag$ & N/S & 2.23 & N/S & 3.06 \\ 
LibriSpeech (noisy)  & 3.59$^\ddag$ & N/S &  3.08$^\ddag$ & N/S & -- & N/S & 2.92 \\ 
Fleurs-Multilingual$_7$  &  6.59$^\ddag$ & N/S  & 4.27$^\ddag$ & N/S & -- & N/S & 5.11 \\ 
\midrule
\multicolumn{5}{l}{\textbf{Speech Translation (BLEU|COMET)}} \\
Fleurs (xx$\rightarrow$en)$_7$              & 32.3|86.9$^\ddag$ &  N/S   & 33.9|87.6$^\ddag$ & N/S & -- & N/S &  34.0|86.9 \\ 
\midrule
\multicolumn{5}{l}{\textbf{Text-to-Speech (WER$\downarrow$)}}  \\
Seed-TTS-Eval (en)                     & -- & N/S  & N/S & N/S & -- & N/S & 1.70 \\ 
\midrule
\multicolumn{5}{l}{\textbf{Audio Generation (OpenL3 Fréchet Distance $\downarrow$)}} \\
AudioCaps                      & N/S & N/S & N/S & N/S & N/S & N/S  & 66.9 \\ 
SongDescriber                     & N/S  & N/S & N/S & N/S & N/S & N/S & 62.7 \\ 
\midrule
\multicolumn{5}{l}{\textbf{Speech-to-Speech}} \\
BigBenchAudio                   &  97.6  & N/S & -- & N/S & 59.0 & N/S & 90.0 \\ 
\bottomrule
\end{tabular}
\end{adjustbox}
\end{table}

We evaluate \model-30B-A3B along two dimensions:
\emph{i)} its text-only capabilities relative to its LLM backbone, including reasoning, knowledge, alignment, long-context understanding, and agentic performance; and
\emph{ii)} its audio-related capabilities, including audio understanding, speech recognition, speech translation, text-to-speech, audio generation, and speech-to-speech generation.
The main results are shown in Table~\ref{tab:main_results}, the detailed results are described in Section \ref{sec:detailed_results}, and Appendix \ref{appendix:benchmarks} includes descriptions on benchmarks and evaluation setups.

As shown in Table~\ref{tab:main_results}, \model achieves accuracies highly comparable to Nemotron-Cascade-2 on text-only benchmarks.
In particular, it achieves even slightly higher accuracies than its text-only backbone on IMO AnswerBench, as well as on AIME 2026 and HMMT Feb2025 with tool-integrated reasoning.
In contrast, reasoning benchmarks show substantial degradation for Qwen3-Omni-30B-A3B-Thinking~\citep{xu2025qwen3omni} compared to its text-only backbone Qwen3-30B-A3B-Thinking-2507.
Qwen3.5-Omni-Flash~\citep{qwen3.5_omni} also performs substantially worse than its LLM counterpart, Qwen3.5.~\footnote{Although Qwen3-Omni-Flash is described in the official report as an in-house variant of Qwen3-Omni-30B-A3B, Qwen3.5-Omni-Flash is proprietary, and the Qwen3.5-Omni technical report does not disclose its relationship to Qwen3.5-35B-A3B.}

On audio understanding benchmarks, \model-30B-A3B achieves performance comparable to the strongest open models, including Step-Audio-R1.1-33B~\citep{tian2025step} and Qwen3-Omni-30B-A3B-Thinking, while only lagging behind the proprietary Qwen3.5-Omni-Flash.
On speech recognition, \model-30B-A3B achieves state-of-the-art performance, outperforming Step-Audio-R1.1-33B and Qwen3-Omni-30B-A3B-Thinking, while performing comparably to the proprietary Qwen3.5-Omni-Flash.
As a versatile model, \model-30B-A3B also delivers compelling performance on speech translation, text-to-speech, and speech-to-speech generation. It is worth noting that \model is the only model among the strongest open models that supports general audio generation beyond speech.

%% file: sections/4_Model_Architecture.tex
\section{Model Architecture}
\label{sec:model_arch}

\begin{figure}[!t]
\centering

\definecolor{panelbg}{HTML}{F5F4F1}   
\definecolor{boxbg}{HTML}{ECEAE4}     
\definecolor{textcol}{HTML}{1A1A1A}   

\resizebox{\textwidth}{!}{%
\begin{tikzpicture}[
    x=1cm,y=1cm,
    text=textcol,
    line cap=round,
    line join=round,
    >={Latex[length=3.0mm,width=2.1mm]},
    box/.style={
        draw=none,
        fill=boxbg,
        rounded corners=2.5pt,
        minimum width=3.65cm,
        minimum height=1.10cm,
        align=center,
        inner sep=2pt,
        font=\bfseries
    },
    smallbox/.style={
        draw=none,
        fill=boxbg,
        rounded corners=2.5pt,
        minimum width=2.3cm,
        minimum height=1.10cm,
        align=center,
        inner sep=2pt,
        font=\bfseries
    },
    dashedbox/.style={
        draw=textcol,
        fill=boxbg,
        dashed,
        dash pattern=on 3.2pt off 3.2pt,
        line width=0.9pt,
        rounded corners=2.5pt,
        minimum width=3.65cm,
        minimum height=1.10cm,
        align=center,
        inner sep=2pt,
        font=\bfseries
    },
    finalbox/.style={
        draw=textcol,
        fill=boxbg,
        line width=1.0pt,
        rounded corners=2.5pt,
        minimum width=3.65cm,
        minimum height=1.10cm,
        align=center,
        inner sep=2pt,
        font=\bfseries
    },
    widebox/.style={
        draw=none,
        fill=boxbg,
        rounded corners=2.5pt,
        minimum width=21.27cm,
        minimum height=1.20cm,
        align=center,
        inner sep=2pt,
        font=\bfseries
    },
    flow/.style={
        draw=textcol,
        line width=0.95pt,
        -{Latex[length=3.0mm,width=2.1mm]}
    }
]

\fill[panelbg, rounded corners=3pt] (0,0) rectangle (23.6,11.2);

\node[anchor=west, font=\bfseries\Large] at (0.62,10.0) {\model Architecture};

\node[anchor=center, widebox] (llm) at (11.8,6.0) {Nemotron-Cascade-2-30B-A3B};  

\node[anchor=center, font=\bfseries] (audioin) at (3.0,0.9) {Speech / General Audio};
\node[box]       (audioenc)  at (3.0,2.4) {Audio Encoder};
\node[box] (mlp) at (3.0,4.2) {MLP Adapters};
\node[anchor=center, font=\bfseries] (textin) at (7.0,4.2) {Text Tokens};

\draw[flow] (audioin.north)  -- (audioenc.south);
\draw[flow] (audioenc.north)  -- (mlp.south);
\draw[flow] (mlp.north)  -- (3.0, 5.4);
\draw[flow] (textin.north)  -- (7.0, 5.4);

\node[anchor=center, font=\bfseries] (textout) at (12.6,7.5) {Text Tokens};
\draw[flow] (12.6, 6.6) -- (textout.south);

\node[anchor=center, font=\bfseries] (speechtok) at (16.6,7.5) {Speech Tokens};
\node[box] (speechdec) at (16.6,9.0) {Speech Decoder};
\node[anchor=center, font=\bfseries] 
(speechout) at (16.6,10.5) {Speech};

\draw[flow] (16.6, 6.6) -- (speechtok.south);
\draw[flow] (speechtok.north) -- (speechdec.south);
\draw[flow] (speechdec.north) -- (speechout.south);

\node[anchor=center, font=\bfseries] (audiotok) at (20.6,7.5) {Audio Tokens};
\node[box] (audiodec) at (20.6,9.0) {Audio Decoder};
\node[anchor=center, font=\bfseries] (audioout) at (20.6,10.5) {General Audio};

\draw[flow] (20.6, 6.6) -- (audiotok.south);
\draw[flow] (audiotok.north) -- (audiodec.south);
\draw[flow] (audiodec.north) -- (audioout.south);

\end{tikzpicture}%
}
\caption{\model-30B-A3B architecture. The LLM backbone is Nemotron-Cascade-2-30B-A3B. Audio inputs are encoded into continuous embeddings using an audio encoder and MLP adapters. \model uses an extended vocabulary and directly predicts discrete speech and audio tokens in addition to text tokens.}

\label{fig:model_arch}
\end{figure}
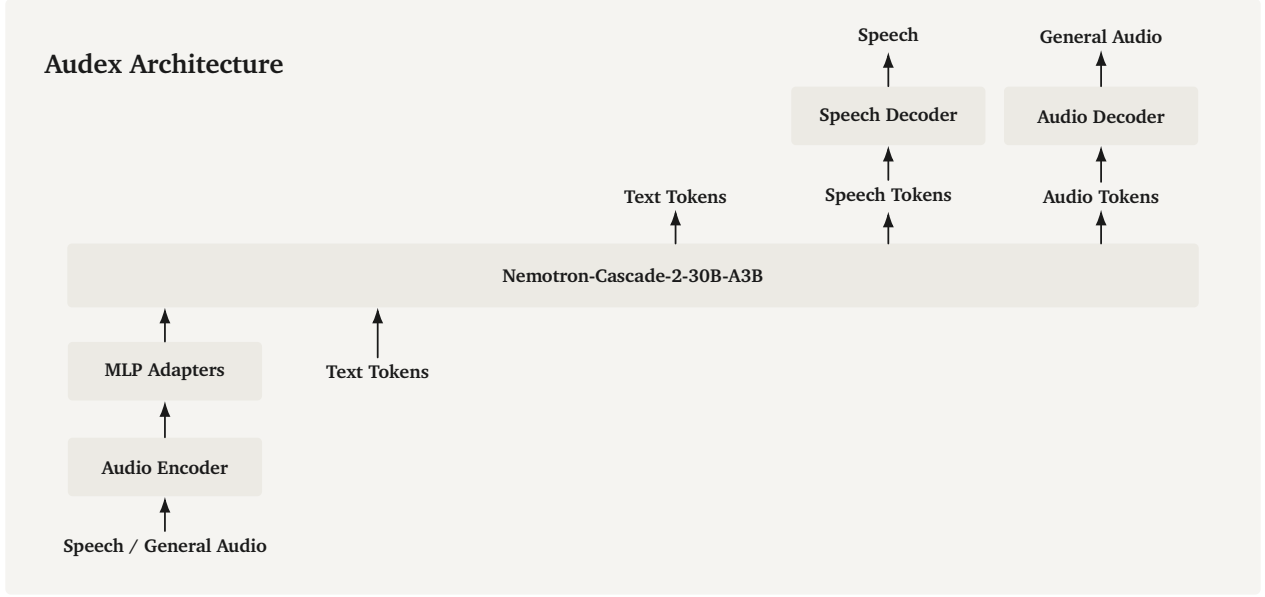

The model architecture of \model is illustrated in Figure~\ref{fig:model_arch}.

\subsection{LLM Backbone}

\model-30B-A3B uses Nemotron-Cascade-2-30B-A3B \citep{Nemotron_Cascade_2} as the backbone LLM. It is a Mixture-of-Experts (MoE) LLM built upon the hybrid Mamba-Transformer~\citep{nemotron_nano_v3} with 52 layers and a model dimension of 2688. There are 128 routable experts and 6 activated experts. The details of the architecture can be found in \citet{nemotron_nano_v3}. 
Specifically, we initialize the weights from the checkpoint obtained after SFT in the post-training process of Nemotron-Cascade-2-30B-A3B~\citep{Nemotron_Cascade_2}.

Furthermore, we build \model-2B with a 2B dense Transformer as the LLM backbone, which has 28 layers, model dimension of 2048, and MLP intermediate size of 9216.
This text-only 2B dense Transformer is trained using the same data and recipes as Nemotron-Cascade-2-30B-A3B. 

The original vocabulary size of both backbone LLMs is 131,072, which we extend to 205,312 to accommodate audio tokens, as described in Section~\ref{sec: model architecture audio codec}. These additional token embeddings are initialized with random Gaussians with mean $0$ and standard deviation $0.02$.

\subsection{Audio Encoder}

We use the finetuned AF-Whisper from Audio Flamingo 3 \citep{goel2025audio} as our audio encoder. AF-Whisper shares the same architecture as Whisper Large-v3 \citep{radford2022whisper} and extends the understanding abilities to general audio tasks beyond speech recognition. The input is 16kHz audio. For every 30 seconds of audio, AF-Whisper outputs audio features at 25Hz (i.e. a length 750 for the 30 seconds) with a dimension of 1280. We split audio into 30-second windows and pad the final partial window to 30 seconds similar to \citet{liu2025voxtral}. We use 2-layer MLP adapters to convert the audio features to the LLM model dimension.

\subsection{Audio Codec}
\label{sec: model architecture audio codec}

Audio codec tokenizes waveform into discrete indices. \model expands its text vocabulary set and assigns the audio codec tokens to the new indices.  \model is then trained to directly predict the audio tokens for audio generation tasks in addition to text tokens using the same cross-entropy loss.

We use different codecs for speech and non-speech generation. Speech is tokenized using a speech-focused codec and non-speech sound is tokenized using a general-purpose audio codec with separate vocabularies. The separate codec dispatch design offers a flexible way to decode audios using specialized strategies for speech and non-speech sound. For example, a speech-focused codec can use a single-layer quantizer with lower tokens per second for efficient prediction, whereas non-speech sound can use a multi-layer quantizer (e.g. residual vector quantization~(RVQ)~\citep{zeghidour2021soundstream}) with higher tokens per second target because non-speech generally has more complex auditory structure that warrants a higher budget.

We use X-Codec2 \citep{ye2025llasa} to tokenize speech. X-Codec2 is a speech-focused codec that operates at 50Hz frame rate and uses a single-layer finite scalar quantization~(FSQ) \citep{mentzer2023finite} with a $65536$ codebook size. That is, speech is tokenized at 50 token/s with $65536$ vocabulary size.

We use X-Codec \citep{ye2025codec} to tokenize non-speech audio, following UALM \citep{tian2025ualm}. X-Codec is a general-purpose audio codec that operates at 50Hz. It uses an 8-layer RVQ~\citep{zeghidour2021soundstream} with $1,024$ codebook size per layer. We use the first 4 RVQ layers to reduce the number of tokens used for \model. We apply a simple flattening pattern \citep{copet2024simple} for the two-dimensional RVQ tokens (frame and depth) so that \model handles the structure of RVQ tokens same as text. Therefore, non-speech is tokenized at $50\times4=200$ token/s with $1024\times4=4096$ vocabulary size. 

The speech and audio token indices are directly appended to the vocabulary as \texttt{<speechcodec\_\{id\}>} and \texttt{<audiocodec\_\{id\}>}. 
We append the full vocabulary per codec: $65536$ speech tokens   and $8192$ non-speech audio tokens. For the latter, we reserve the full $8\times1024=8192$ entries corresponding to the 8-layer RVQ depth, even though only the first 4 layers are used for \model. 
We also append special tokens for audio generation: \texttt{<speechgen\_start>}, \texttt{<speechgen\_end>}, \texttt{<audiogen\_start>}, and \texttt{<audiogen\_end>} are inserted at the start and end of generation, along with a placeholder \texttt{<audiogen>} token that reserves a region for generation replaced by \texttt{<speechcodec\_\{id\}>} or  \texttt{<audiocodec\_\{id\}>} for a given training sample using the audio codec. This results in a total of  $131072+65536+8192+5=204805$ vocabulary size for \model. To support tensor parallelism in Megatron-LM framework \citep{shoeybi2019megatron}, we further pad the model's embedding table to size $205312$, which is divisible by $512$.

The inputs and outputs of the audio codecs are in 16kHz. We apply two strategies for audio decoding: $(i)$~for non-speech audio from X-Codec, to improve perceptual audio quality and bandwidth, we train a separate enhancement VAE as in \citet{tian2025ualm} with an acoustic reconstruction training recipe based on BigVGAN-v2 \citep{lee2023bigvgan}. It increases the sampling rate to 48kHz and reduces codec artifacts. $(ii)$ For faster inference from the predicted X-Codec2 tokens, we train a streaming decoder using a causal ConvNeXt architecture \citep{liu2022convnet} based on Vocos \citep{siuzdak2023vocos} to replace X-Codec2's non-causal decoder. We find using a small look-ahead context window ensures high semantic accuracy and maintains fast streaming inference.

%% file: sections/5_Training_Recipes.tex
\section{Model Training}
\label{sec:training}

\subsection{Training Data}

\model is trained on a mix of text, audio, and interleaved text-audio data. The text-only data are similar to those in LLM post-training. The audio data include speech data and general non-speech sound. For speech, we train on classic automatic speech recognition (ASR), automatic speech translation (AST), and text-to-speech (TTS) datasets. For general non-speech sound, we train on audio understanding (audio and text in, text out) and text-to-audio (TTA) datasets. The dataset statistics can be found in Table \ref{tab:data_statistics}.

We follow Nemotron-Cascade-2 \citep{Nemotron_Cascade_2} in terms of text supervised fine-tuning~(SFT) and reinforcement learning~(RL) data. The dataset includes a wide range of tasks for reasoning, math, coding, alignment, long-context tasks, agentic tasks, and multi-lingual tasks from Nemotron-Cascade~\citep{Nemotron_Cascade_1} and Nemotron 3 Nano~\citep{nemotron_nano_v3}. 
For text SFT data, we filter samples that exceed the total length of 200K tokens. 

For ASR and AST, we use publicly available speech corpora with permissive licenses.  
We format each ASR target response as the input language followed by the transcript.
To improve ASR robustness across diverse noisy and reverberant environments, we augment the dataset by generating noisy and reverberant versions of its clean speech. We use real full-band noise and room impulse responses from the 5th DNS Challenge resources~\citep{dubey2024icassp}, and keep the original transcript as the target.
For AST, each target response contains the source language, the source-language transcription, and the English translation in sequence, allowing the model to learn speech translation through an intermediate transcription step.

For TTS, we use publicly available speech corpora with permissive licenses, including dedicated TTS corpora and clean samples from the ASR data.
For fixed-voice TTS, we select a synthetic voice from Tortoise \citep{betker2023tortoise} that has good reconstruction quality with X-Codec 2, and use Qwen3-TTS-12Hz-1.7B-Base \citep{Qwen3-TTS} to convert about 8\% of the TTS data to this voice. 
We threshold input audio lengths to be between 0.1 and 120 seconds; nevertheless, most of the samples are between a few and twenty seconds.

We follow Audio Flamingo 3 \citep{goel2025audio} and Audio Flamingo Next \citep{ghosh2026audio} in terms of audio understanding data. We use samples in the format of single-turn audio question-answering and re-sample all audio to 16kHz during pre-processing. We threshold input audio lengths to be between 0.1 and 900 seconds.

We follow the methods in ETTA \citep{lee2024etta}, Stable Audio Open \citep{evans2024stable}, Tango-AF \citep{kong2024improving}, and TangoFlux \citep{hung2024tangoflux} to curate our text-to-audio generation data. We use permissively licensed samples from these datasets, and apply filters to remove low-quality or uncorrelated samples. We fix output length to be 10 seconds, as training on variable lengths may introduce instability in early experiments.

\begin{table}[!h]
    \setlength{\cmidrulewidth}{\heavyrulewidth}
    \centering
    \caption{\model data statistics grouped by data types and tasks. The number of samples refer to the number of input--output pairs for text and audio tasks, where partial duplications are considered as different samples (e.g. different inputs for the same output, or different outputs for the same inputs). The total audio hours are measured under 16kHz. The total audio tokens are measured with X-Codec (RVQ4) for general audio and X-Codec2 for speech. The total text tokens are measured with the \model tokenizer. The total tokens is a simple addition of audio and text tokens.}
    \label{tab:data_statistics}
    \begin{TableSmall}
    \begin{adjustbox}{max width=\textwidth}
    \begin{tabular}{lrrrrr}
    \toprule
        Task & \# samples & audio hours & audio tokens & text tokens & total tokens \\ \cmidrule(lr){1-1} \cmidrule(lr){2-6}
        Text-to-Speech (TTS)              & 147M & 421K    & 75.8B  &  5.2B & 81.0B  \\
        Text-to-Audio  (TTA)              &  12M &  34K  & 24.3B  &  0.4B & 24.7B  \\
        Audio Understanding         &  49M & 308K  & 27.7B  &  2.2B & 29.9B  \\
        Speech Recognition (ASR)                         &  95M & 201K  & 18.1B  &  4.4B & 22.5B  \\
        Speech Translation (AST)         &  58M & 128K  & 11.5B  &  5.3B & 16.8B  \\
        Text SFT                    &  33M &      0  &     0  & 303.0B & 303.0B \\
        \cmidrule(lr){1-1} \cmidrule(lr){2-6}
        \textbf{Total} & \textbf{394M} & \textbf{1{,}092K} & \textbf{157.4B} & \textbf{320.5B} & \textbf{477.9B} \\
    \bottomrule
    \end{tabular}
    \end{adjustbox}
    \end{TableSmall}
\end{table}

\subsection{Chat Templates}
We include all of our chat templates in Figure \ref{fig: templates}. The text chat and tool-calling templates follow \citet{Nemotron_Cascade_2}. The audio understanding, speech recognition, and speech translation are all formatted via the audio QA format, where the input audio is represented by the \texttt{<sound>} token, the user asks a question, and the assistant provides the answer in the non-thinking mode. For text-to-speech generation, the user turn is fixed to ``\textit{\texttt{<|text to speech|>} Generate speech for this transcription. \{transcription\}}'', and the model outputs the speech tokens. Similarly, for text-to-audio generation, the user turn is fixed to ``\textit{\texttt{<|text to audio|>} Generate audio for this caption. \{caption\}}'', and the model outputs the audio tokens.

\newcommand{\chattemplatebox}[3]{%
\definecolor{panelbg}{HTML}{F5F4F1}
\definecolor{textcol}{HTML}{1A1A1A}
\begin{tikzpicture}
  \node[
    fill=panelbg,
    text=textcol,
    rounded corners=4pt,
    inner sep=12pt,
    text width=0.98\linewidth,
    align=left,
    minimum height=4cm
  ] {%
    \begin{minipage}[t]{\linewidth}
      {\bfseries\normalsize #1}\par
      {\color{textcol}\rule{#3\linewidth}{0.9pt}}\par
      \vspace{0.5em}
      {\footnotesize #2}
    \end{minipage}
  };
\end{tikzpicture}%
}

\begin{figure}[htbp]
\centering

\begin{subfigure}[T]{0.46\textwidth}
\centering
\chattemplatebox{Chat Templates}{
\textbf{<im\_start>system}\\
You are a helpful and harmless assistant.\\
You are not allowed to use any tools\textbf{<|im\_end|>}\\
\textbf{<im\_start>user}\\
User contents. \textbf{<im\_end>}\\
\textbf{<im\_start>assistant}\\
\textit{<think>}\\
Model thinking contents.\\
\textit{</think>}\\
Model outputs.\textbf{<im\_end>}
}{0.5}
\caption{Chat Templates for Text (same as \citet{Nemotron_Cascade_2}). To use the non-thinking mode, we directly append the \textit{</think>} token after \textit{<think>} without "$\backslash$n".}
\end{subfigure}
\hfill
\begin{subfigure}[T]{0.46\textwidth}
\centering
\chattemplatebox{Audio QA Templates}{
\textbf{<im\_start>system}\\
You are a helpful and harmless assistant.\\
You are not allowed to use any tools\textbf{<|im\_end|>}\\
\textbf{<im\_start>user}\\
User contents (e.g., Transcribe the speech in the input audio; Translate the spoken content in the audio to English; Where is the communication likely taking place?). 
\textbf{<sound>}\textbf{<im\_end>}\\
\textbf{<im\_start>assistant}\\
\textit{<think>}\textit{</think>}\\
Model outputs.\textbf{<im\_end>}
}{0.6}
\caption{Audio Question-Answering Templates. The \textbf{<sound>} special token will be replaced by sound embeddings \textbf{<so\_start>}<so\_embedding>...<so\_embedding>\textbf{<so\_end>} computed from the MLP adapters.}
\end{subfigure}

\vspace{0.6cm}

\begin{subfigure}[T]{0.46\textwidth}
\centering
\chattemplatebox{Text-to-Speech Templates}{
\textbf{<im\_start>system}\\
You are a helpful and harmless assistant.\\
You are not allowed to use any tools\textbf{<|im\_end|>}\\
\textbf{<im\_start>user}\\
\textbf{<|text to speech|>} Generate speech for this transcription. Hello, how are you today?\textbf{<im\_end>}\\
\textbf{<im\_start>assistant}\\
\textit{<think>}\textit{</think>}\\
\textbf{<speechgen\_start>}<speechcodec\_123>...\\<speechcodec\_456>\textbf{<speechgen\_end>}\textbf{<im\_end>}
}{0.75}
\caption{Text-to-Speech Templates. We use the \textbf{<audiogen>} placeholders in data manifests and replace these with the actual speech codec tokens after loading the \texttt{base64}-decoded representations (see Section \ref{sec: training_recipes_infra_training}).}
\end{subfigure}
\hfill
\begin{subfigure}[T]{0.46\textwidth}
\centering
\chattemplatebox{Text-to-Audio Templates}{
\textbf{<im\_start>system}\\
You are a helpful and harmless assistant.\\
You are not allowed to use any tools\textbf{<|im\_end|>}\\
\textbf{<im\_start>user}\\
\textbf{<|text to audio|>} Generate audio for this caption. Birds chirping in a forest.\textbf{<im\_end>}\\
\textbf{<im\_start>assistant}\\
\textit{<think>}\textit{</think>}\\
\textbf{<audiogen\_start>}<audiocodec\_123>...\\<audiocodec\_456>\textbf{<audiogen\_end>}\textbf{<im\_end>}
}{0.75}
\caption{Text-to-Audio Templates. We use the \textbf{<audiogen>} placeholders in data manifests and replace these with the actual audio codec tokens after loading the \texttt{base64}-decoded representations (see Section \ref{sec: training_recipes_infra_training}).}
\end{subfigure}

\caption{Templates for (a) text-only chats (see tool-calling templates in \citet{Nemotron_Cascade_2}), (b) audio question-answering, which includes audio understanding, speech recognition, and speech translation, (c) text-to-speech synthesis, and (d) text-to-audio synthesis.}
\label{fig: templates}
\end{figure}

\subsection{Training Loss and Classifier-Free Guidance}

\model is trained with the standard LLM loss function:

\begin{equation}
\label{eq: llm}
    \max_{\theta}~~\left\{\frac{1}{\text{\#Trainable tokens}}\sum_{i:~x_i\text{ is trainable}}\log p_{\theta}(x_i|x_{<i})\right\},
\end{equation}
where $x_i$ is the $i$-th token in a sample, and $p_{\theta}(x_i|x_{<i})$ is the model's predicted likelihood of $x_i$ given all prior tokens $\{x_1,x_2,\cdots,x_{i-1}\}$. In a packed sequence of $k$ samples, let $L_j$ be the vector of Cross-Entropy losses of the $j$-th sample, and $|L_j|$ the length of $L_j$. We use the square averaging loss function \citep{chen2024expanding} \footnote{\url{https://github.com/NVIDIA/Megatron-LM/blob/main/examples/multimodal/train.py}}
  
\begin{equation}
    \label{eq: scaled loss}
    \mathrm{Scaled\text{-}Loss}(\{L_1, \cdots, L_k\}) = \left(\sum_{j=1}^k\frac{\mathrm{sum}(L_j)}{\sqrt{|L_j|}}\right) \cdot \left(\sum_{j=1}^k\sqrt{|L_j|}\right)^{-1}.
\end{equation}

For the \model-30B-A3B model MoE layers, we use  auxiliary-loss-free load balancing \citep{wang2024auxiliary, liu2024deepseek} with an update rate of $10^{-3}$, top-K routing scale of $2.5$, together with the standard load balancing loss \citep{lepikhin2020gshard} with a coefficient of $10^{-6}$.

To enable classifier-free guidance (CFG) \citep{ho2022classifier}, we need to learn an unconditional distribution $p_{\theta}(\cdot|\emptyset)$. This will enable CFG sampling at inference time by modifying the logits to:
\begin{equation}
\label{eq: cfg}
    p_{\theta}^{\mathrm{CFG}}(\cdot|x_{<i})) = p_{\theta}(\cdot|x_{<i})) + (\lambda-1)\cdot \left[p_{\theta}(\cdot|x_{<i}))-p_{\theta}(\cdot|\emptyset))\right].
\end{equation}
Similar to UALM \citep{tian2025ualm}, we set $\emptyset$ to be $i-1$ consecutive, un-trainable padding tokens so that it has the same lengths as $x_{<i}$. In practice, the common way to enable CFG training is to randomly flip the conditions $x_{<i}$ to $\emptyset$ at at fixed probability (e.g., 10\%) \citep{ho2022classifier}. Since flipping the tokens on-the-fly could complicate our training infrastructure and parallelism implementation, and we only need CFG training for certain tasks (speech and audio generation), we use an equivalent pre-processing approach. Specifically, we randomly select 10\% of speech and audio generation data, convert their transcriptions or captions into a random number of padding tokens, and blend these data into the training dataset.

\subsection{Training Stages}

\subsubsection{Supervised Fine-Tuning}
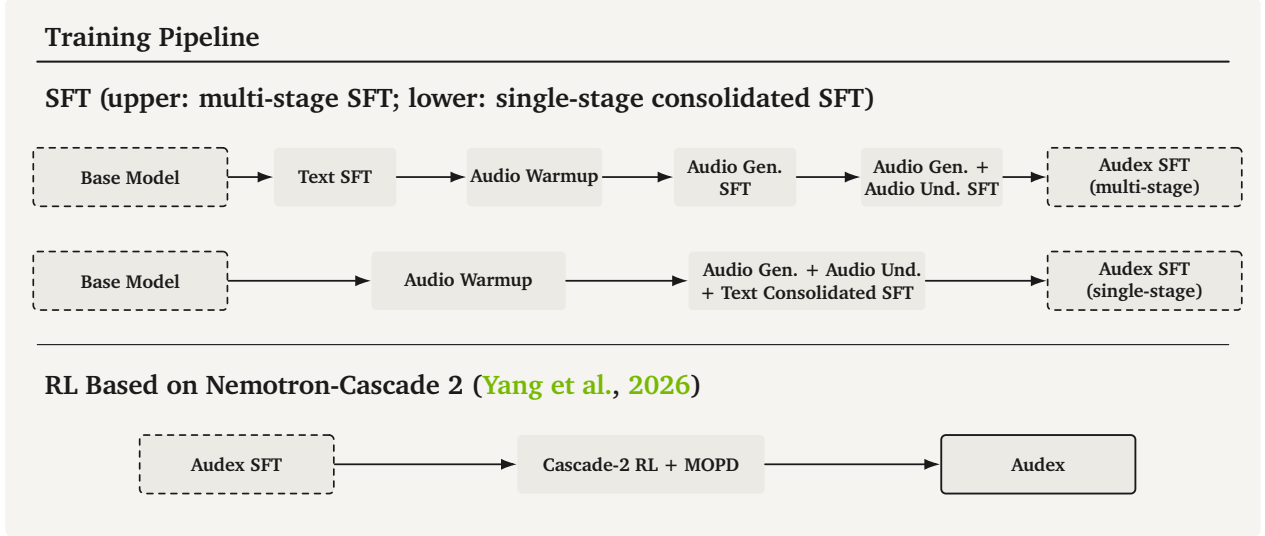
\begin{figure}[!t]
\centering

\definecolor{panelbg}{HTML}{F5F4F1}   
\definecolor{boxbg}{HTML}{ECEAE4}     
\definecolor{textcol}{HTML}{1A1A1A}   

\resizebox{\textwidth}{!}{%
\begin{tikzpicture}[
    x=1cm,y=1cm,
    text=textcol,
    line cap=round,
    line join=round,
    >={Latex[length=3.0mm,width=2.1mm]},
    box/.style={
        draw=none,
        fill=boxbg,
        rounded corners=2.5pt,
        minimum width=3.65cm,
        minimum height=1.10cm,
        align=center,
        inner sep=2pt,
        font=\bfseries
    },
    smallbox/.style={
        draw=none,
        fill=boxbg,
        rounded corners=2.5pt,
        minimum width=2.3cm,
        minimum height=1.10cm,
        align=center,
        inner sep=2pt,
        font=\bfseries
    },
    dashedbox/.style={
        draw=textcol,
        fill=boxbg,
        dashed,
        dash pattern=on 3.2pt off 3.2pt,
        line width=0.9pt,
        rounded corners=2.5pt,
        minimum width=3.65cm,
        minimum height=1.10cm,
        align=center,
        inner sep=2pt,
        font=\bfseries
    },
    finalbox/.style={
        draw=textcol,
        fill=boxbg,
        line width=1.0pt,
        rounded corners=2.5pt,
        minimum width=3.65cm,
        minimum height=1.10cm,
        align=center,
        inner sep=2pt,
        font=\bfseries
    },
    flow/.style={
        draw=textcol,
        line width=0.95pt,
        -{Latex[length=3.0mm,width=2.1mm]}
    }
]

\fill[panelbg, rounded corners=3pt] (0,0) rectangle (23.6,10.1);

\node[anchor=west, font=\bfseries\Large] at (0.62,9.35) {Training Pipeline};
\draw[textcol, line width=1.1pt] (0.6,8.92) -- (23.0,8.92);


\node[anchor=west, font=\bfseries\Large] at (0.62,8.2) {SFT (upper: multi-stage SFT; lower: single-stage consolidated SFT)};

\node[dashedbox] (base1)    at  (2.35,6.75) {Base Model};
\node[smallbox](textsft1)   at  (6.20,6.75) {Text SFT};
\node[smallbox](warmup1)    at  (9.95,6.75) {Audio Warmup};
\node[smallbox](gen1)       at (13.72,6.75) {Audio Gen.\\SFT};
\node[smallbox](genund1)    at (17.42,6.75) {Audio Gen. +\\Audio Und. SFT};
\node[dashedbox](nasftmultistage) at (21.42,6.75) {\model SFT\\(multi-stage)};

\draw[flow] (base1.east) -- (textsft1.west);
\draw[flow] (textsft1.east) -- (warmup1.west);
\draw[flow] (warmup1.east) -- (gen1.west);
\draw[flow] (gen1.east) -- (genund1.west);
\draw[flow] (genund1.east) -- (nasftmultistage.west);

\node[dashedbox] (base2)  at  (2.35,4.8) {Base Model};
\node[box](warmup2)  at (8.71,4.8) {Audio Warmup};
\node[box](fusion)  at (15.07,4.8) {~~Audio Gen. + Audio Und.\\+ Text Consolidated SFT};
\node[dashedbox](nasftfusion) at (21.42,4.8) {\model SFT\\(single-stage)};

\draw[flow] (base2.east) -- (warmup2.west);
\draw[flow] (warmup2.east) -- (fusion.west);
\draw[flow] (fusion.east) -- (nasftfusion.west);

\draw[textcol, line width=0.6pt] (0.6,3.6) -- (23.0,3.6);

\node[anchor=west, align=center, font=\bfseries\Large] at (0.62,2.8) {RL Based on Nemotron-Cascade 2 \citep{Nemotron_Cascade_2}};

\node[dashedbox]      (nasft)   at (4.35,1.35) {\model SFT};
\node[box](cascade)  at (11.95,1.35) {~~~~Cascade-2 RL + MOPD~~~~};
\node[finalbox] (final) at (19.42,1.35) {\model};

\draw[flow] (nasft.east)  -- (cascade.west);
\draw[flow] (cascade.east)  -- (final.west);

\end{tikzpicture}%
}
\caption{\model training stages. In the SFT stage, we study two training curriculums: (1) the multi-stage curriculum -- which adds one type of capabilities at a time, and (2) the consolidated single-stage curriculum -- which combines all types of task together. The resulting model is called \model SFT. We then apply cascaded RL in the text domain similar to Nemotron-Cascade 2, and obtain \model.}
\label{fig: training pipeline}

\end{figure}

In this work, we find that it is possible to build a best-in-class audio LLM without audio- or speech-related pretraining.
We start with the text-only pretrained LLM and study two post-training strategies (see Figure \ref{fig: training pipeline} (upper)). The first is the multi-stage SFT strategy, where we decouple different tasks in different stages and gradually add more tasks. 

\begin{itemize}
    \item We first apply text-only SFT same as Nemotron-Cascade 2 SFT \citep{Nemotron_Cascade_2}. Indeed, we directly reuse the checkpoint after SFT from  Nemotron-Cascade 2.
    \item We then extend the vocabulary to include audio and speech tokens, add the audio encoder and initialized MLP adapters, and warmup these components. We call this the audio warmup stage; in this stage, we only train on the MLP adapters and audio token embeddings. We freeze text token embeddings because training on these leads to degraded text quality (see analysis in Appendix \ref{app: warmup}). 
    \item After audio warmup, we conduct audio generation SFT. We unfreeze the entire LLM and train on audio/speech generation data plus text-only data to maintain the text abilities.
    We introduce this stage without audio-to-text tasks as audio and speech token generation are entirely new tasks for the model, and thus require more training to learn effectively.
    Note that audio encoder and MLP are not involved with audio/speech generation and text-only data.
    \item Finally, we further unfreeze the MLP adapters and the LLM, and include audio understanding, ASR, and AST tasks, along with audio generation and text-only data. The audio encoder remains frozen.
\end{itemize}

The second SFT strategy is the single-stage consolidated SFT. This is inspired by recent studies \citep{team2024chameleon,team2026kimi,tong2026beyond} which state that it is helpful to include multimodal data in early training stages. In detail, we directly apply audio warmup to the pretrained base LLM, and then combine all SFT data (audio generation, audio understanding, and text) for a single SFT stage. \footnote{In both SFT strategies, we have applied an extra stage to remove speech prompting abilities for responsible model release.}

In Table \ref{tab: SFT data blend}, we demonstrate the data blending ratios (upper) and key hyper-parameters (lower) at different SFT stages. The data blending ratios are shown in epochs for clarity. 

\begin{table}[!h]
    \setlength{\cmidrulewidth}{\heavyrulewidth}
    \centering
    \caption{Data blending weights (upper) and key training hyper-parameters at different SFT stages.}
    \begin{adjustbox}{width={1.0\textwidth}}
    \begin{tabular}{lcccccc}
    \toprule
    \multicolumn{7}{c}{\multirow{2}{*}{\textbf{Data Blending Ratios (Epochs by Data Types)}}} \\ & \\ \cmidrule{1-7}
    \multirow{2}{*}{Training Data Type}
    & \multicolumn{4}{c}{Multi-stage SFT}
    & \multicolumn{2}{c}{Single-stage SFT} \\
    \cmidrule(lr){2-5} \cmidrule(lr){6-7}
    & Text
    & Audio Warmup
    & Audio Gen.
    & Audio Gen. + Audio Und.
    & Audio Warmup
    & Consolidated SFT \\
    \cmidrule(lr){1-1} \cmidrule(lr){2-5} \cmidrule(lr){6-7}
    Text                & 1.6 & 0.5 & 0.5 & 1.5 & 0.5 & 2.0 \\
    ASR+AST             & 0.0 & 0.0 & 0.0 & 1.0 & 0.0 & 1.0 \\
    Audio understanding & 0.0 & 2.0 & 0.0 & 2.0 & 2.0 & 2.0 \\
    Text-to-speech      & 0.0 & 1.0 & 1.0 & 1.0 & 1.0 & 1.0 \\
    Text-to-audio       & 0.0 & 2.0 & 1.0 & 1.0 & 2.0 & 1.0 \\
    \hline 
    Text weights (\%)   & 100 & 44 & 59 & 69 & 44 & 75 \\ \cmidrule{1-7}
    \multicolumn{7}{c}{\multirow{2}{*}{\textbf{Training Hyper-parameters}}} \\ & \\ \cmidrule{1-7}
    \multirow{2}{*}{Parameter}
    & \multicolumn{4}{c}{Multi-stage SFT}
    & \multicolumn{2}{c}{Single-stage SFT} \\
    \cmidrule(lr){2-5} \cmidrule(lr){6-7}
    & Text
    & Audio Warmup
    & Audio Gen.
    & Audio Gen. + Audio Und.
    & Audio Warmup
    & Consolidated SFT \\
    \cmidrule(lr){1-1} \cmidrule(lr){2-5} \cmidrule(lr){6-7}
    Global Batch Size & 64 & 64 & 16 & 64 & 64 & 64 \\
    Maximum Learning Rate & $5\times10^{-5}$ & $2\times10^{-3}$ & $2\times10^{-5}$ & $2\times10^{-5}$ & $2\times10^{-3}$ & $2\times10^{-5}$ \\
    Minimum Learning Rate & $5\times10^{-6}$ & $1\times10^{-5}$ & $1\times10^{-6}$ & $1\times10^{-6}$ & $1\times10^{-5}$ & $1\times10^{-6}$ \\
    Learning Rate Warmup & 200 steps & 5\% & 5\% & 5\% & 5\% & 5\% \\
    Learning Rate Scheduler & cosine & cosine & cosine & cosine & cosine & cosine \\
    AdamW $\beta_1,\beta_2$ & 0.9, 0.98 & 0.9, 0.999 & 0.9, 0.999 & 0.9, 0.999 & 0.9, 0.999 & 0.9, 0.999 \\
    Weight Decay & 0.1 & 0.0 & 0.05 & 0.05 & 0.0 & 0.05 \\
    \multirow{2}{*}{Trainable Modules} & \multirow{2}{*}{LLM} & MLP Adapters + & \multirow{2}{*}{LLM} & MLP Adapters + & MLP Adapters + & MLP Adapters + \\
    & & Audio Token Emb. & & LLM & Audio Token Emb. & LLM \\
    \bottomrule
    \end{tabular}
    \end{adjustbox}
    
    \label{tab: SFT data blend}
\end{table}

The comparison between the multi-stage and single-stage SFT is shown in Section \ref{sec: multi-stage vs single-stage SFT}. While the results of these two models are comparable overall, the most significant difference is that the single-stage SFT model has much worse quality on NIAH, indicating that the attention mechanism has gone through drastic changes during the consolidated SFT stage.

\subsubsection{Comparison between multi-stage and single-stage SFT}
\label{sec: multi-stage vs single-stage SFT}

We compare the 30B-A3B checkpoints after multi-stage and single-stage SFT on representative benchmarks in Table~\ref{tab: multi-stage vs single-stage}. Overall, the two checkpoints achieve comparable performance, with the exception that the single-stage SFT checkpoint fails on long-context tasks (NIAH) almost completely. We hypothesize that this is because long text samples and long audio samples are too different and difficult for the base LLM to learn simultaneously, while the multi-stage SFT handles this by having good long-context text abilities first and maintaining this afterwards. 
Besides, we find the single-stage SFT checkpoint is slightly better on speech understanding and recognition tasks, but slightly worse on audio generation. Nevertheless, none of the rest of the results indicate a clear gap between these two checkpoints. 

By simply adding the epochs across training stages, we find the multi-stage SFT requires more training budget than the consolidated single-stage SFT. In summary, while the single-stage SFT can achieve similar results at lower total training budget, we favor the multi-stage SFT strategy due to the following reasons: $(i)$ the single-stage SFT breaks the long-context attention mechanism, $(ii)$ the multi-stage SFT is easy to tune (stage-by-stage), while the single-SFT is much harder to stabilize (see Appendix \ref{app: text blending ratios} for analysis).

\begin{table}[t!]
\centering
\footnotesize
\renewcommand{\arraystretch}{1.25}
\caption{Multi-Stage vs. Single-Stage SFT for \model-30B-A3B SFT. We mark tool-integrated reasoning results with $\diamondsuit$ for math and code reasoning.}
\begin{TableSmall}
    \begin{adjustbox}{max width=\textwidth}
    \begin{tabular}{l|cc}
    \toprule
    \textbf{Benchmark} & \textbf{Multi-Stage SFT} & \textbf{Single-Stage Consolidated SFT} \\
    \midrule
    \multicolumn{3}{l}{\textbf{Reasoning}} \\
    AIME 2025 & 91.7|96.3$^\diamondsuit$ & 90.5|97.2$^\diamondsuit$ \\
    HMMT Feb25 & 92.2|91.4$^\diamondsuit$ & 87.1|93.1$^\diamondsuit$ \\
    LiveCodeBench v6  & 82.3|85.3$^\diamondsuit$ & 80.1|83.6$^\diamondsuit$ \\
    \midrule
    \multicolumn{3}{l}{\textbf{Knowledge}} \\
    MMLU-Redux         & 86.3 & 86.8 \\
    MMLU-Pro           & 77.5 & 78.4 \\
    GPQA-Diamond       & 73.8 & 72.2 \\
    \midrule
    \multicolumn{3}{l}{\textbf{Alignment}} \\
    ArenaHard v2      & 59.4 & 61.4 \\
    IFBench (prompt)  & 52.4 & 52.4 \\
    \midrule
    \multicolumn{3}{l}{\textbf{Long Context }} \\
    AA-LCR                  & 27.0 & 29.8 \\
    NIAH (256K|1M) & 99.3|86.8 & 6.0|0.0 \\
    \midrule
    \multicolumn{3}{l}{\textbf{Agentic}} \\
    $\tau^2$-Bench              & 52.4 & 52.4 \\
    \midrule
    \midrule
    \multicolumn{3}{l}{\textbf{Audio Understanding}} \\
    MMAU                   & 75.5 & 75.4 \\ 
    MMAR                    & 62.7 & 62.5 \\ 
    MMSU                    & 63.3 & 63.5 \\ 
    Audio Entailment$_2$ & 95.3 & 94.9 \\
    CMM Hallucination & 87.1 & 83.8 \\
    \midrule
    \multicolumn{3}{l}{\textbf{Speech Recognition (WER$\downarrow$)}} \\ 
    LibriSpeech (clean|other)  & 1.33|2.99 & 1.32|2.98 \\ 
    LibriSpeech (noisy)  & 2.75 & 2.71 \\ 
    Fleurs-Multilingual$_7$  & 5.07 & 4.79  \\ 
    \midrule
    \multicolumn{3}{l}{\textbf{Speech Translation (BLEU|COMET)}} \\
    Fleurs (xx$\rightarrow$en)$_7$              & 34.1|86.9 & 34.4|87.0 \\ 
    \midrule
    \multicolumn{3}{l}{\textbf{Text-to-Speech (WER$\downarrow$)}}  \\
    Seed-TTS-Eval (en)                     & 1.76 & 1.75 \\ 
    \midrule
    \multicolumn{3}{l}{\textbf{Audio Generation (OpenL3 Fréchet Distance $\downarrow$)}} \\
    AudioCaps                      & 60.9 & 69.4 \\ 
    SongDescriber                     & 58.1 & 64.4 \\ 
    \midrule
    \multicolumn{3}{l}{\textbf{Speech-to-Speech}} \\
    BigBenchAudio                   & 88.0 & 84.6  \\ 
    \bottomrule
    \end{tabular}
    \end{adjustbox}
    \end{TableSmall}
\label{tab: multi-stage vs single-stage}
\end{table}

\subsubsection{Cascade RL and Multi-Domain On-policy Distillation}
We therefore select the multi-stage SFT checkpoint as our \model SFT checkpoint, and proceed with RL training (see Figure \ref{fig: training pipeline} (lower)). 
In this work, we only do text-only RL. 
The RL stage is very similar to Nemotron-Cascade~2 \citep{Nemotron_Cascade_2}: we begin with Instruction-Following RL, then Multi-domain RL, Multi-domain On-policy Distillation~(MOPD), Reinforcement Learning from Human Feedback~(RLHF), and finish with Long-context RL. 
Interestingly, the audio and speech-related tasks show marginal or no regression after text-only RL, while the model significantly improves on text tasks.
The underlying reason is well explained in Section~4.1.1 of Nemotron-Cascade~\citep{Nemotron_Cascade_1}, \emph{Why Cascade RL for LLMs Is Resistant to Catastrophic Forgetting.}
This sheds light on a promising direction for further improving audio and speech-related tasks through audio-text RL, without causing regression on text benchmarks.
We will pursue it as future work.
The final checkpoint is our \model-30B-A3B model.~\footnote{We only apply RL to the 30B-A3B model. The 2B model is the multi-stage SFT checkpoint.}

\subsection{Infrastructure}

\paragraph{Training: Megatron Core}
\label{sec: training_recipes_infra_training}
We use Megatron framework \citep{shoeybi2019megatron}\footnote{\url{https://github.com/NVIDIA/Megatron-LM}} with Transformer Engine\footnote{\url{https://github.com/NVIDIA/TransformerEngine}} to train \model. We use 4-way tensor parallelism (TP), 32-way expert parallelism (EP), 8-way context parallelism (CP), and sequence parallelism (SP) to scale the model training using $262,144$ context length. We train \model using 512 NVIDIA H100 GPUs with BF16 precision and the AdamW optimizer, with gradient clipping set to $1.0$.

\paragraph{Data Loading: Megatron Energon}
We use Megatron-Energon\footnote{\url{https://github.com/NVIDIA/Megatron-Energon}} multimodal dataloader and an online sequence packing using a balanced greedy knapsack algorithm, same as in Nemotron 3 Nano Omni \citep{deshmukh2026nemotron}. We use $8,192$ packing buffer size and \texttt{num\_workers=4} per data-parallel (DP) rank. To minimize training overhead from an audio codec encoder forward pass to extract target tokens, we use pretokenized audio samples for TTS and TTA data under the \texttt{base64} format. 

\paragraph{Classifier-Free Guidance (CFG)}
For CFG training, we pre-select unconditional generation samples with their text replaced by $\emptyset$ (padding tokens with a randomized length) and blend them together with the original dataset.
At inference time, we support CFG sampling in vLLM by submitting each generation as a paired conditional/unconditional sample in the same decoding batch. A custom logits processor tracks the pairs, combines their next-token logits using \eqref{eq: cfg}, and feeds them back into sampling. The pairs are kept token-synchronized after each sampling step.

%% file: sections/6_Results.tex
\section{Detailed Results and Analysis}
\label{sec:detailed_results}

In this section, we present the main results of \model-30B-A3B and \model-2B. Section \ref{sec: main text} includes the text benchmark results. Section \ref{sec: main tts tta} includes the text-to-speech and text-to-audio benchmark results. Section \ref{sec: main audio qa} includes the ASR, AST, and audio understanding results. Section \ref{sec: main s2s} includes the speech interaction benchmark results. The details on benchmarks and evaluation setups are in Appendix \ref{appendix:benchmarks}. The detailed results across SFT and RL training stages are presented in Appendix \ref{app: results by stages}. Extra ablation studies are presented in Appendix \ref{app: ablation}. 

\subsection{Results on Text Benchmarks}
\label{sec: main text}

We report the text benchmark results for \model-30B-A3B, \model-2B and other state-of-the-art baseline models in Table~\ref{tab: main text results}.
The detailed evaluation setups for text benchmarks are described in Appendix~\ref{app: benchmark reasoning} -- \ref{app: benchmark agent}.
For the baseline models, we use numbers from the official report when available; otherwise, we evaluate them using the recommended settings, as indicated by $^\ddag$.

\model-30B-A3B achieves best-in-class results, substantially outperforming leading open and proprietary models of comparable size on reasoning and alignment benchmarks. It also achieves comparable performance on knowledge, long-context, and agentic benchmarks, with results that are sometimes higher and sometimes lower than the strongest baselines.

\begin{table}[!h]
\centering
\footnotesize
\renewcommand{\arraystretch}{1.15}
\caption{Results on text benchmarks. 
We mark tool-integrated reasoning results with $\diamondsuit$ for math and code reasoning.}
\label{tab: main text results}
\begin{TableSmall}
    \begin{adjustbox}{max width=\textwidth}
    \begin{tabular}{lccccc|>{\columncolor{c3light}}c>{\columncolor{c3light}}c}
    \toprule
    \shortstack[l]{\textbf{Benchmark} \\ \textbf{} \\ \textbf{}}
     & \shortstack{\textbf{Step-Audio}\\ \textbf{R1.1} \\ \textbf{33B}}
     & \shortstack{\textbf{Voxtral}\\ \textbf{Small-24B} \\ \textbf{2507}}
     & \shortstack{\textbf{MiMo-Audio}\\ \textbf{7B} \\ \textbf{}}
     & \shortstack{\textbf{Qwen3-Omni}\\ \textbf{30B-A3B} \\ \textbf{Thinking}}
     & \shortstack{\textbf{Qwen3.5-Omni}\\ \textbf{Flash} \\ \textbf{35B-A3B}}
     & \shortstack{\textbf{\model} \\ \textbf{30B-A3B} \\ \textbf{}}
     & \shortstack{\textbf{\model} \\ \textbf{2B} \\ \textbf{}} \\
    \midrule
    Context Length & 64K & 128K & 8K & 64K & 256K & 1M & 128K \\ 
    \midrule
    \multicolumn{5}{l}{\textbf{Reasoning}} \\
    AIME 2025 & 44.8$^\ddag$ & -- & -- & 73.7 & -- & 91.2|98.3$^\diamondsuit$ & 68.9|68.3$^\diamondsuit$ \\
    AIME 2026 & 57.8$^\ddag$ & -- & -- & -- & -- & 89.4|96.6$^\diamondsuit$ & -- \\
    HMMT Feb25 & 27.0$^\ddag$ & -- & -- & 60.4$^\ddag$ & 59.0 & 92.2|93.8$^\diamondsuit$  & 68.5|67.3$^\diamondsuit$ \\
    IMO AnswerBench & -- & -- & -- & 59.9$^\ddag$ & 51.5 & 81.1 & -- \\
    LiveCodeBench v6  & 22.9$^\ddag$ & 22.9$^\ddag$ & -- & 59.2$^\ddag$ & 56.6 & 85.3|86.2$^\diamondsuit$ & 64.5|72.7$^\diamondsuit$ \\
    SciCode            & -- & -- & -- & 30.6 & 25.5 & 35.2 & -- \\
    \midrule
    \multicolumn{5}{l}{\textbf{Knowledge}} \\
    MMLU-Redux         & 86.6$^\ddag$ & 74.5$^\ddag$ & 73.8$^\ddag$ & 88.8 & 90.0 & 86.4 & 70.2 \\
    MMLU-Pro           & 75.3$^\ddag$ & 60.5$^\ddag$ & 55.3$^\ddag$ & 80.4$^\ddag$  & 79.9 & 78.9 & 57.0 \\
    GPQA-Diamond       & 60.7$^\ddag$ & 37.9$^\ddag$ & 13.8$^\ddag$ & 73.1 & 76.4 & 74.9 & 47.2 \\
    \midrule
    \multicolumn{5}{l}{\textbf{Alignment}} \\
    ArenaHard v2      & 44.3$^\ddag$ & 5.0$^\ddag$ & 18.7$^\ddag$ & 55.1 & -- & 81.6 & 11.9 \\
    IFBench (prompt)  & 32.2$^\ddag$ & 25.2$^\ddag$ & 24.4$^\ddag$ & 52.4$^\ddag$ & 38.4 & 77.8 & 42.6 \\
    \midrule
    \multicolumn{5}{l}{\textbf{Long Context }} \\
    AA-LCR                  & -- & -- & -- & 11.0$^\ddag$ & 46.0 & 39.3 & 17.4 \\
    LongBench v2            & -- & -- & -- & 27.2$^\ddag$ & 46.4 & 41.3 & -- \\
    NIAH (256K|1M) & -- & -- & -- & 0.0|0.0$^\ddag$ & -- & 99.4|83.4 & -- \\
    \midrule
    \multicolumn{5}{l}{\textbf{Agentic}} \\
    $\tau^2$-Bench              & -- & -- & -- & 45.4 & 78.0 & 57.2 & 41.7 \\
    Terminal Bench 2.0          & -- & -- & -- & 1.1$^\ddag$ & -- & 19.1 & -- \\
    SWE Verified       & -- & -- & -- & 23.4$^\ddag$ & -- & 48.2 & -- \\ 
    \bottomrule
    \end{tabular}
    \end{adjustbox}
    \end{TableSmall}
\end{table}

\subsection{Results on Speech and Audio Generation}
\label{sec: main tts tta}

\paragraph{Text-to-Speech Results}
The TTS evaluation results on Seed-TTS-Eval (English subset) are shown in Table \ref{tab: main TTS}. The prompting word-error-rate (WER) and speaker similarity (SIM) are measured with the \model-SFT checkpoint, as the prompting function is erased and we only keep fixed-voice TTS afterwards. Results show that \model achieves a very low WER, despite that the SIM is lower, which is possibly because our codec choice and training recipes do not optimize for this. 

\begin{table}[!t]

    \centering
    \caption{TTS results on Seed-TTS-Eval (English). We measure both prompting and fixed-voice TTS.}
    \begin{TableSmall}
    \begin{adjustbox}{max width=\textwidth}
        \begin{tabular}{l|cc|c}
            \toprule
            Models & WER (prompting) $\downarrow$ & SIM & WER (fixed voice) $\downarrow$ \\
            \hline
            Seed-TTS & 2.25 & 76.2 & -- \\
            Step-Audio-TTS-3B & 2.31 & 66.0 & -- \\
            E2 TTS & 2.19 & 71.0 & -- \\
            F5 TTS & 1.83 & 64.7 & -- \\
            Qwen2.5-Omni-7B & 2.33 & 64.1 & -- \\
            Qwen3-Omni-30B-A3B-Instruct & -- & -- & 1.39 \\
            MiMo-Audio & -- & -- & 5.37 \\
            \hline
            \rowcolor{c3light} \model 30B-A3B & 2.07 & 45.3 & 1.70 \\
            \rowcolor{c3light} \model 2B & -- & -- & 2.04 \\
            \bottomrule
        \end{tabular}
    \end{adjustbox}
    \end{TableSmall}
    \label{tab: main TTS}
\end{table}

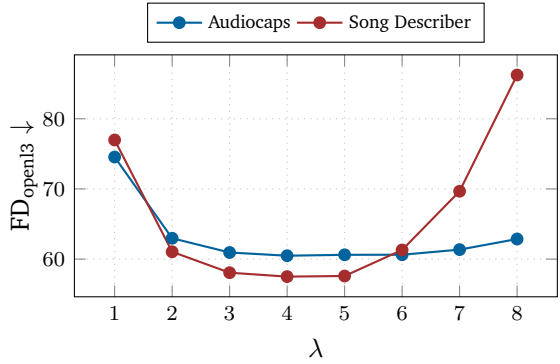
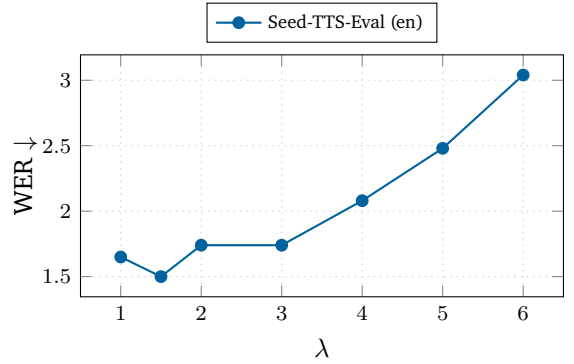
\begin{figure}[!t] 
    \centering
    \begin{subfigure}[t]{0.48\textwidth}
        \centering
        \captionsetup{width=0.9\linewidth}
        \begin{tikzpicture}
        \begin{axis}[
            width=\textwidth,
            height=0.6\textwidth,
            xlabel={$\lambda$},
            xtick={1,2,3,4,5,6,7,8},
            ylabel={FD$_\text{openl3}$ $\downarrow$},
            ylabel style={yshift=-0.5em},
            grid=both,
            grid style={dotted, gray!50},
            mark options={solid},
            ticklabel style={font=\footnotesize},
            legend style={font=\scriptsize, at={(0.5,1.05)}, anchor=south, legend columns=2},
        ]
        \addplot[color=c0, thick, mark=*] coordinates {
            (1, 74.55) (2, 62.97) (3, 60.94) (4, 60.48)
            (5, 60.61) (6, 60.62) (7, 61.35) (8, 62.86)
        };
        \addlegendentry{Audiocaps}
        
        \addplot[color=c5, thick, mark=*] coordinates {
            (1, 76.98) (2, 61.04) (3, 58.06) (4, 57.50)
            (5, 57.59) (6, 61.30) (7, 69.67) (8, 86.24)
        };
        \addlegendentry{Song Describer}
        \end{axis}
        \end{tikzpicture}
        \caption{FD$_\text{openl3}$ ($\downarrow$) vs CFG value $\lambda$ on TTA benchmakrs.}
    \end{subfigure}
    \hfill
    \begin{subfigure}[t]{0.48\textwidth}
        \centering
        \captionsetup{width=0.9\linewidth}
        \begin{tikzpicture}
        \begin{axis}[
            width=\textwidth,
            height=0.6\textwidth,
            xlabel={$\lambda$},
            xtick={1,2,3,4,5,6},
            ylabel={WER $\downarrow$},
            ylabel style={yshift=-0.5em},
            grid=both,
            grid style={dotted, gray!50},
            mark options={solid},
            ticklabel style={font=\footnotesize},
            legend style={font=\scriptsize, at={(0.5,1.05)}, anchor=south, legend columns=2},
        ]
        \addplot[color=c0, thick, mark=*] coordinates {
            (1, 1.65) 
            (1.5, 1.5) 
            (2, 1.74)
            (3, 1.74)
            (4, 2.08)
            (5, 2.48)
            (6, 3.04)
        };
        \addlegendentry{Seed-TTS-Eval (en)}
        
        \end{axis}
        \end{tikzpicture}
        \caption{WER ($\downarrow$) vs CFG value $\lambda$ on the Seed-TTS-Eval benchmark.}
    \end{subfigure}

    \caption{The Effect of CFG Value $\lambda$ on TTA and TTS. CFG is required to achieve good TTA quality, and the optimal range is $\lambda\in[3,5]$. CFG is not necessary for good TTS quality, although using $\lambda\approx1.5$ may lead to slightly better WER than not using CFG while doubling the inference budget.}
    \label{fig: CFG for audio gen}
\end{figure}

\paragraph{Text-to-Audio Results}
The TTA evaluation results on AudioCaps \citep{kim2019audiocaps} and SongDescriber \citep{manco2023thesong} are shown in Table \ref{tab: main TTA}. We measure $\mathrm{FD}_{\mathrm{openl3}}$, the Fréchet Distance using the OpenL3 \citep{cramer2019look, arandjelovic2017look} feature space as suggested by \citet{evans2024stable}. $\mathrm{FD}_{\mathrm{openl3}}$ measures the overall distributional distance between generated audio and the test set and is the lower the better. We use the enhancement VAE as described in Section \ref{sec: model architecture audio codec}, and use the best inference hyper-parameters as shown in Figure \ref{fig: CFG for audio gen}.

\model 30B-A3B, as a versatile audio language model, matches strong diffusion and autoregressive task-specific baselines. Notably, the \model 30B-A3B SFT checkpoint achieves 60.9 on AudioCaps and 58.1 on SongDescriber (see Appendix \ref{app: results by stages}), which are the state-of-the-art results among models trained on public data. After our text RL, we find there is marginal drop of the scores ($+6.0/4.6$, respectively), which is negligible for $\mathrm{FD}_{\mathrm{openl3}}$. In addition, the \model 2B model also achieves strong results on these two benchmarks. 

Most diffusion and autoregressive baselines are conditioned on semantic text embedding via cross-attentions. These could make the model easier to understand the semantics and therefore easier to learn TTA. In contrast, our \model directly places the raw text tokens in the user turn similar to a chat format, and generates raw audio tokens. With proper scaling and CFG training (which was first discovered in our previous work, UALM \citep{tian2025ualm}), we unify TTA with the LLM's next-token-prediction paradigm, and enable future directions such as fine-grained controllable generation and reasoning across modalities. 

Compared to the task-specific baselines that can generate long and consistent outputs, we find \model works the best for a fixed 10-second duration. We tried training on variable lengths in early experiments but found that makes training less stable and outputs to be more random (both in terms of duration and contents). This is possibly because most of the training data are fixed to be 10 seconds. We aim to solve the duration and consistency issues with further high-quality data curation, filtering, scaling, and RL.

\begin{table}[!t]
    \centering
    \caption{TTA Results on AudioCaps and SongDescriber Using $\mathrm{FD}_{\mathrm{openl3}}~(\downarrow)$. For both test sets, the models are evaluated on 10-second segments of the full test set. The model type indicates whether the model is a diffusion model, an autoregressive model, or an actual autoregressive LLM with strong text abilities. The text prompt conditioning colum displays how the TTA model is conditioned on the input text prompt. $\spadesuit$: uses non-public, licensed training corpora.}
    \begin{TableSmall}
    \begin{adjustbox}{max width=\textwidth}
        \begin{tabular}{l|rr|cc}
            \toprule
            Models & Model Type & Text Prompt Conditioning & AudioCaps & SongDescriber \\
            \hline
            AudioLDM2 Large & Diffusion & Flan-T5 embeddings & 121.6 & 331.7 \\
            Stable Audio Open & Diffusion & T5/Flan-T5 embeddings &  100.9 & 138.6 \\
            ETTA & Diffusion & T5 embeddings & 80.1 & 95.7 \\
            Tango2 & Diffusion & Flan-T5 embeddings & 108.4 & -- \\
            TangoFlux & Diffusion & Flan-T5 embeddings & 103.0 & 235.6 \\
            MusicGen‑Stereo‑L$^{\spadesuit}$ & Autoregressive & T5 embeddings & -- & 228.9 \\
            Magenta RealTime$^{\spadesuit}$ & Autoregressive & MusicCoCa embeddings & -- & 40.5 \\
            UALM-Gen & Autoregressive & BPE Tokens & 75.1 & 74.4 \\
            UALM & LLM & BPE Tokens & 65.9 & 83.7 \\
            \hline
            \rowcolor{c3light} \model 30B-A3B & LLM & BPE Tokens & 66.9 & 62.7 \\
            \rowcolor{c3light} \model 2B & LLM & BPE Tokens & 79.3 & 78.4 \\
            \bottomrule
        \end{tabular}
    \end{adjustbox}
    \end{TableSmall}
    \label{tab: main TTA}

\end{table}

\paragraph{Inference Hyper-parameters for Speech and Audio Generation}
The inference hyper-parameters for TTS and TTA are very unique and different from the typical values in text reasoning. In Figure \ref{fig: CFG for audio gen}, we study how the CFG value $\lambda$ in \eqref{eq: cfg} affects generation quality. \footnote{$\lambda=1$ means no CFG.} We find for TTA, $\lambda\in[3,5]$ leads to the best overall quality, and therefore select $\lambda=3$ in our experiments. For TTS, $\lambda\approx1.5$ leads to the best WER, and therefore select $\lambda=1.5$ in our experiments. Large $\lambda$ values lead to worse quality as expected. However, small $\lambda$ values affect TTA more significantly than TTS -- even without CFG, the TTS WER is low, indicating that \model can be used in TTS applications without sacrificing inference speed for quality. 

We follow UALM and use Top-k sampling. In Appendix \ref{app: cfg}, we study the best temperature and Top-k value. We find the best Top-k value is 80, and the best temperature is 1.0 for TTA and 0.1 for TTS. This suggests that TTA requires more randomness at inference time, while TTS does not require much randomness.

\subsection{Results on Audio Understanding, ASR, and AST}
\label{sec: main audio qa}

In this section, we evaluate \model on audio understanding, ASR, and AST tasks. All of these are formatted in the audio question-answering template. For audio understanding tasks, the user question is defined in the specific benchmark, and we use a top-p of 0.9 and temperature of 0.7. For ASR and AST tasks, we use fixed prompts in the user turn, and use greedy sampling.

\paragraph{English ASR Results}
The evaluation results on OpenASR leaderboard \citep{srivastav2025openasr} \footnote{\url{https://huggingface.co/spaces/hf-audio/open_asr_leaderboard}} are shown in Table \ref{tab: open asr leaderboard}. 
The results on noisy LibriSpeech ASR evaluation following \citep{nvidia2025canary1bv2} are shown in Table \ref{tab: noisy asr}. 
For both benchmarks, we report word error rate (WER) represented in percentage (the lower the better).
Results show that \model matches strong audio-language model baselines and outperforms several strong ASR-specific baselines, especially in the noisy ASR setting. The gap between our 30B-A3B and 2B models is not huge, as the audio encoder perceives most of the information needed for transcribing.

\begin{table}[!t]
    \centering
    \caption{WER ($\downarrow$) Evaluation Results on OpenASR Leaderboard (English).}
    \begin{TableSmall}
    \begin{adjustbox}{max width=\textwidth}
    \begin{tabular}{l|cccccccc|c}
    \toprule
        Models & LS-clean & LS-other & AMI & Earnings22 & GigaSpeech & SPGI Speech & TED-LIUM & VoxPopuli & Average \\
        \midrule

        Whisper-large-v3 & 2.01 & 3.91 & 15.95 & 11.29 & 10.02 & 2.94 & 3.86 & 9.54 & 7.44 \\
        Canary-1B-v2 & 2.18 & 3.56 & 16.01 & 11.79 & 10.82 & 2.28 & 4.29 & 6.25 & 7.15 \\
        Parakeet-TDT-0.6B-v3 & 1.92 & 3.59 & 11.39 & 11.19 & 9.57 & 3.98 & 2.80 & 6.09 & 6.32 \\
        Canary Qwen 2.5B & 1.61 & 3.10 & 10.19 & 10.45 & 9.43 & 1.90 & 2.71 & 5.66 & 5.63 \\
        
        Step-Audio-R1.1 33B 
        & 1.66
        & 3.28
        & 22.36
        & 11.54
        & 9.97
        & 3.80
        & 3.03
        & 7.60
        & 7.91 \\ 
        
        Qwen3-Omni-30B-A3B-Instruct 
        & 1.28
        & 2.52
        & 11.44
        & 10.47
        & 8.79
        & 2.44
        & 2.78
        & 6.07
        & 5.72 \\

        Qwen3-Omni-30B-A3B-Thinking
        & 1.77
        & 3.70
        & 25.19
        & 10.51
        & 9.49
        & 3.47
        & 2.38
        & 7.52
        & 8.00 \\

        Qwen3.5-Omni Flash & 1.30 & 2.43 & -- & -- & -- & -- & -- & -- & -- \\

        Kimi-Audio-Instruct 
        & 1.28
        & 2.42
        & 18.19
        & 12.25
        & 10.25
        & 3.53
        & 2.90
        & 8.06
        & 6.62 \\

        Voxtral-Small-24B-2507
        & 1.59
        & 3.26
        & 15.27
        & 10.50
        & 9.81
        & 2.02
        & 3.52
        & 6.96
        & 6.62 \\

        MiMo-Audio & 3.50 & -- & -- & -- & -- & -- & -- & -- & -- \\

        \midrule
        \rowcolor{c3light} \model 30B-A3B & 1.34 & 3.06 & 17.24 & 11.92 & 9.90 & 1.76 & 3.50 & 5.83 & 6.82 \\
        
        \rowcolor{c3light} \model 2B 
        & 1.63
        & 3.77
        & 17.10
        & 11.95
        & 10.86
        & 2.00
        & 3.64
        & 6.17
        & 7.14 \\ 
    \bottomrule
    \end{tabular}
    \end{adjustbox}
    \end{TableSmall}
    \label{tab: open asr leaderboard}
\end{table}

\begin{table}[!t]
    \centering
    \caption{WER ($\downarrow$) Evaluation Results on Noisy Librispeech ASR at different noise levels.}
    \begin{TableSmall}
    \begin{adjustbox}{max width=\textwidth}
    \begin{tabular}{l|ccccccc|c}
    \toprule
        Models & -5dB & 0dB & 5dB & 10dB & 25dB & 50dB & 100dB  & Average \\
        \midrule
        Whisper-large-v3 
        & 9.30
        & 4.24
        & 2.72
        & 2.27
        & 2.15
        & 1.98
        & 1.97 
        & 3.52 \\
        Canary-1B-v2 
        & 19.38
        & 5.08
        & 2.80
        & 2.29
        & 2.01
        & 2.16
        & 2.18 
        & 5.13 \\
        Parakeet-TDT-0.6B-v3 
        & 12.21
        & 4.82
        & 2.62
        & 2.15
        & 1.96
        & 1.92
        & 1.92 
        & 3.94 \\
        Canary Qwen 2.5B
        & 11.06
        & 4.73
        & 2.50
        & 2.91
        & 1.64
        & 1.61
        & 1.61 
        & 3.58 \\
        Step-Audio-R1.1 33B  
        & 9.74 
        & 4.05
        & 2.74
        & 3.55
        & 1.72
        & 1.69
        & 1.61 
        & 3.59 \\
        Qwen3-Omni-30B-A3B-Instruct 
        & 7.43
        & 2.75
        & 1.74
        & 1.49
        & 1.34
        & 1.28
        & 1.26 
        & 2.47 \\
        Qwen3-Omni-30B-A3B-Thinking
        & 8.63
        & 3.66
        & 2.29
        & 1.95
        & 1.72
        & 1.69
        & 1.64 
        & 3.08 \\
        Kimi-Audio-Instruct 
        & 9.81
        & 3.86
        & 2.06
        & 1.60
        & 1.47
        & 1.35
        & 1.33 
        & 3.07 \\
        Voxtral-Small-24B-2507
        & 8.39
        & 3.68
        & 2.20
        & 1.87
        & 1.61
        & 1.58
        & 1.59 
        & 2.99 \\

        \midrule
        \rowcolor{c3light} \model 30B-A3B 
        & 9.37
        & 3.26
        & 2.01
        & 1.67
        & 1.41
        & 1.37
        & 1.34 
        & 2.92 \\
        \rowcolor{c3light} \model 2B & 10.60 & 4.02 & 2.41 & 1.97 & 1.66 & 1.60 & 1.63 & 3.41 \\ 
    \bottomrule
    \end{tabular}
    \end{adjustbox}
    \end{TableSmall}
    \label{tab: noisy asr}
\end{table}

\paragraph{Multi-Lingual ASR and AST Results}
We evaluate multi-lingual ASR and AST using the Fleurs benchmark \citep{conneau2023fleurs}, and evaluate on seven languages: 
DE	(German),
ES	(Spanish),
FR	(French),
IT	(Italian),
PT	(Portuguese),
RU	(Russian),
KO	(Korean). 
We report the WER for multi-lingual ASR in Table \ref{tab: multilingual asr}, and BLEU \citep{papineni2002bleu} and COMET~\citep{rei2020comet} scores for AST (the higher the better) in Table \ref{tab: ast}. Results show that \model also has strong multi-lingual ASR and AST abilities that are comparable to several state-of-the-art speech LLMs. \footnote{We report the average of these seven (noted as Fleurs-Multilingual$_7$ and Fleurs (xx$\rightarrow$en)$_7$) in Table \ref{tab:main_results}.}

\begin{table}[!t]
    \centering
    \caption{WER ($\downarrow$) Evaluation Results on Fleurs Multilingual ASR.}
    \begin{TableSmall}
    \begin{adjustbox}{max width=\textwidth}
    \begin{tabular}{l|ccccccc|c}
    \toprule
        Models & DE & ES & FR & IT & PT & RU & KO & Average\\
        \midrule
        Whisper-large-v3     
            & 3.28
            & 2.33
            & 4.99
            & 2.41
            & 3.50
            & 3.64
            & 7.09
            & 3.89 \\
        Seamless-m4t-v2-large
            & 5.29
            & 3.99
            & 6.37
            & 3.52
            & 6.94
            & 6.51
            & 14.76
            & 6.77 \\
        Canary-1B-v2
            & 3.47
            & 2.63
            & 4.41
            & 2.35
            & 4.19
            & 5.92
            & --
            & -- \\
        Parakeet-TDT-0.6B-v3 
            & 4.27
            & 3.28
            & 4.70
            & 2.44
            & 4.54
            & 4.93
            & --
            & -- \\
        Step-Audio-R1.1 33B 
            & 5.49
            & 3.74
            & 6.72
            & 3.25
            & 5.36
            & 9.84
            & 11.75
            & 6.59 \\
        Voxtral-Small-24B-2507 
            & 2.58
            & 2.77
            & 3.54
            & 2.16
            & 3.71
            & 6.29
            & 17.10
            & 5.45 \\
        
        Qwen3-Omni-30B-A3B-Instruct 
            & 2.47
            & 2.34
            & 3.35
            & 1.55
            & 3.07
            & 3.15
            & 19.85
            & 5.11 \\
        Qwen3-Omni-30B-A3B-Thinking
            & 3.40
            & 2.71
            & 4.46
            & 2.52
            & 3.73
            & 5.81
            & 7.29
            & 4.27 \\
        
        \midrule
        \rowcolor{c3light} \model 30B-A3B 
            & 4.33
            & 2.85
            & 5.79
            & 3.27
            & 4.35
            & 5.45
            & 9.71
            & 5.11 \\
        \rowcolor{c3light} \model 2B 
            & 5.98
            & 3.71
            & 6.89
            & 4.22
            & 4.93
            & 8.74
            & 10.22
            & 6.38 \\
    \bottomrule
    \end{tabular}
    \end{adjustbox}
    \end{TableSmall}
    \label{tab: multilingual asr}
\end{table}

\begin{table}[!t]
    \centering
    \caption{Evaluation Results on Fleurs AST (xx$\rightarrow$en) Using BLEU|COMET ($\uparrow$).}
    \begin{TableSmall}
    \begin{adjustbox}{max width=\textwidth}
    \begin{tabular}{l|ccccccc|c}
    \toprule
        Models & DE & ES & FR & IT & PT & RU & KO & Average \\
        \midrule
        Whisper-large-v3      
            & 33.6|84.8
            & 22.5|83.5
            & 31.1|84.4
            & 23.0|84.0
            & 36.8|85.8
            & 26.5|83.3
            & 19.2|81.7
            & 27.5|83.9 \\
        Seamless-m4t-v2-large
            & 36.8|86.0
            & 25.7|84.2
            & 33.9|85.2
            & 26.8|85.0
            & 37.4|85.0
            & 30.2|83.9
            & 22.8|83.5
            & 30.5|84.7 \\
        Canary-1B-v2
            & 35.2|85.4
            & 25.2|83.3
            & 33.7|85.2
            & 25.1|84.2
            & 38.6|85.4
            & 27.0|81.6
            & --
            & -- \\
        Step-Audio-R1.1 33B 
            & 38.0|88.2
            & 26.8|86.4
            & 36.2|87.3
            & 26.7|86.7
            & 41.7|87.7
            & 30.2|85.1
            & 26.4|86.8
            & 32.3|86.9 \\
            
        Voxtral-Small-24B-2507 
            & 43.9|89.0
            & 34.8|87.5
            & 42.6|88.6
            & 35.2|88.0
            & 48.7|89.2
            & 37.2|86.6
            & 26.6|85.7
            & 41.0|88.5 \\
            
        Qwen3-Omni-30B-A3B-Instruct 
            & 41.5|88.5
            & 28.6|86.7
            & 38.1|87.9
            & 28.6|87.0
            & 43.3|88.3
            & 32.4|86.0
            & 28.1|87.6
            & 34.4|87.4 \\
        Qwen3-Omni-30B-A3B-Thinking
            & 40.7|88.7
            & 28.0|86.6
            & 37.6|88.3
            & 28.2|87.0
            & 42.7|88.5
            & 32.3|86.1
            & 27.6|87.9
            & 33.9|87.6 \\
        \midrule
        \rowcolor{c3light} \model 30B-A3B 
            & 40.0|88.0
            & 28.5|86.1
            & 38.0|87.1
            & 29.0|86.8
            & 43.4|88.0
            & 32.8|85.3
            & 26.3|86.9
            & 34.0|86.9\\
        \rowcolor{c3light} \model 2B 
            & 37.3|86.8
            & 35.2|85.8
            & 26.0|85.6
            & 26.5|85.3
            & 41.3|87.0
            & 29.4|83.5
            & 23.4|85.5
            & 31.2|85.7 \\ 
    \bottomrule
    \end{tabular}
    \end{adjustbox}
    \end{TableSmall}
    \label{tab: ast}
\end{table}

\paragraph{Audio Understanding Results}
We evaluate audio understanding quality with the following benchmarks: 
    
\begin{itemize}
    \item MMAU (v05.15.25) \citep{sakshi2024mmau}, MMAR \citep{ma2025mmar}, and MMSU \citep{wang2025mmsu} for general sound, speech, and music understanding. These test samples are all multiple choice questions.
    
    \item Audio Entailment \citep{audioentail} on Clotho v2 and AudioCaps for deductive reasoning. \footnote{We report the average of these two (noted as Audio Entailment$_2$) in Table \ref{tab:main_results}.} These benchmarks contain closed-ended questions and require outputs to be within \{yes, no, maybe\}.
    
    \item CMM \citep{leng2026curse} for hallucination detection. This benchmark also contains closed-ended questions with a focus on sound event existence, and requires outputs to be within \{yes, no\}.
\end{itemize}

The results are shown in Table \ref{tab: audio understanding}. \model matches strong baselines on MMAU, but has some gap on MMAR and MMSU compared to the state-of-the-art audio LLMs. \model demonstrates strong results on audio entailment and CMM hallucination, which indicate better trustworthiness.

\begin{table}[!t]
    \centering
    \caption{Audio Understanding Evaluation Results.}
    \begin{TableSmall}
    \begin{adjustbox}{max width=\textwidth}
    \begin{tabular}{l|ccccc|cc|cc|c}
    \toprule
        \multirow{2}{*}{Models} & \multicolumn{5}{c|}{MMAU} &  \multirow{2}{*}{MMAR} & \multirow{2}{*}{MMSU} & \multicolumn{2}{c|}{Audio Entailment} & \multirow{2}{*}{CMM} \\
        & mini & sound & music & speech & avg &  &  & Clotho v2 & AudioCaps &  \\
        \midrule
        Audio-Flamingo-3 & 74.3 & 75.3 & 74.6 & 69.6 & 73.2 & 60.1 & 62.3 & 93.3 & 95.0 & 86.5 \\
        Step-Audio-R1.1 33B & 76.8 & 76.4 & 68.9 & 75.6 & 73.6 & 69.8 & 74.1 & 59.3 & 63.7 & 83.3 \\ 
        Kimi-Audio-Instruct & 68.2 & 70.7 & 65.9 & 56.6 & 64.4 & 57.4 & 59.3 & 62.4 & 65.3 & 67.0 \\
        Voxtral-Small-24B-2507 & 60.2 & 54.3 & 51.1 & 66.7 & 57.4 & 32.8 & 56.6 & 55.1 & 57.2 & 72.4 \\
        MiMo-Audio & 74.7 & 77.2 & 69.7 & 70.8 & 72.6 & 63.6 & 62.9 & 71.7 & 78.5 & 82.3 \\
        Qwen3-Omni-30B-A3B-Instruct & -- & -- & -- & -- &  77.5 & 67.9 & 69.0 & 66.8 & 68.8 & 90.4 \\
        Qwen3-Omni-30B-A3B-Thinking & -- & -- & -- & -- & 75.4 & 66.4 & 70.2 & 60.4 & 62.8 & 85.0 \\
        \midrule
        
        \rowcolor{c3light} \model 30B-A3B & 76.5 & 81.5 & 77.5 & 67.9 & 75.6 & 63.2 & 63.4 & 94.4 & 95.6 & 90.3 \\
        \rowcolor{c3light} \model 2B & 72.1 & 75.1 & 72.3 & 61.8 & 69.7 & 56.8 & 57.6 & 93.2 & 92.9 & 75.9 \\ 
    \bottomrule
    \end{tabular}
    \end{adjustbox}
    \end{TableSmall}
    \label{tab: audio understanding}
\end{table}

\subsection{Results on Speech Interaction}
\label{sec: main s2s}

We evaluate speech interaction (speech-to-text and speech-to-speech) results via a cascaded system (ASR + text reasoning + potential TTS), but different from most cascaded systems, we use the same \model model with suitable inference setups for all of these three steps. The results on BigBenchAudio are shown in Table \ref{tab: bigbenchaudio}. \model 30B-A3B achieves a score of 90, which is very competitive on the benchmark. \model 2B also outperforms several strong baselines that are 7B or larger. The results on VoiceBench are shown in Table \ref{tab: voicebench}. \model is competitive especially on instruction following, reasoning, multiple choice QA, and safety.

\begin{table}[!h]
    \centering
    \caption{Evaluation Results on BigBenchAudio.}
    \begin{TableSmall}
    \begin{adjustbox}{max width=\textwidth}
    \begin{tabular}{l|c}
    \toprule
        Models & Accuracies \\
        \midrule
        Fun-Realtime-Audiochat & 97.6 \\
        Step-Audio R1.1 Realtime & 97.6 \\
        Gemini 3.1 Flash Live Preview - High & 96.6 \\
        Grok Voice Agent & 93.3 \\
        Gemini 2.5 Flash Native Audio Dialog Thinking & 90.7 \\
        Nova 2.0 Sonic (Mar 2026) & 88.1 \\
        GPT Realtime & 83.3 \\
        GPT-Realtime-1.5 & 81.4 \\
        Qwen3.5 Omni Plus Realtime & 73.0 \\
        Gemini 3.1 Flash Live Preview - Minimal & 71.3 \\
        GPT-4o mini Realtime (Dec 2024) & 68.9 \\
        Gemini 2.5 Flash Native Audio Dialog & 68.6 \\
        GPT Realtime Mini (Oct 2025) & 63.6 \\
        MiMo-Audio & 60.2 \\
        Qwen3.5 Omni Flash Realtime & 59.0 \\
        Qwen3 Omni Flash & 58.7 \\
        Qwen3 Omni Realtime & 56.8 \\
        GPT-4o audio chatcompletions & 54.3 \\
        Kimi-Audio-Instruct & 51.0 \\
        Nemotron VoiceChat & 38.8 \\
        Freeze-Omni & 33.4 \\
        PersonaPlex & 19.1 \\
        FLM-Audio & 16.0 \\
        Moshi & 4.4 \\

        \midrule
        \rowcolor{c3light} \model 30B-A3B & 90.0 \\
        \rowcolor{c3light} \model 2B & 64.3 \\ 
    \bottomrule
    \end{tabular}
    \end{adjustbox}
    \end{TableSmall}
    \label{tab: bigbenchaudio}
\end{table}

\begin{table}[!h]
    \centering
    \caption{Evaluation Results on VoiceBench.}
    \begin{TableSmall}
    \begin{adjustbox}{max width=\textwidth}
    \begin{tabular}{l|ccccccccc}
    \toprule
        Models  & AlpacaEval & CommonEval & WildVoice & SD-QA & MMSU & OpenBookQA & BBH & IFEval & AdvBench \\
        \midrule
        Qwen3-Omni 30B-A3B Instruct & 
        94.8 & 90.8 & 91.6 & 76.9 & 68.1 & 89.7 & 80.4 & 77.8 & 99.3 \\ 
        Qwen3-Omni 30B-A3B Thinking & 
        96.4 & 90.5 & 90.5 & 78.1 & 83.0 & 94.3 & 88.9 & 80.6 & 97.2 \\
        Ultravox-GLM-4P7 & 97.4 & 86.0 & 91.0 & 84.2 & 83.8 & 94.7 & 87.2 & 76.3 & 99.2 \\
        GPT-4o-Audio & 95.6 & 89.8 & 91.6 & 75.5 & 80.2 & 89.2 & 84.1 & 76.0 & 98.7 \\
        Kimi-Audio-Instruct & 89.2 & 79.4 & 84.0 & 63.1 & 62.2 & 83.5 & 69.7 & 61.1 & 100.0 \\
        MiMo-Audio & 75.0 & 64.6 & 63.5 & 36.3 & 52.8 & 66.2 & 68.4 & 69.5 & 98.3 \\
        Baichuan-Audio & 88.2 & 81.6 & 78.4 & 45.8 & 53.2 & 71.7 & 54.8 & 50.3 & 99.4 \\
        Phi-4-multimodal & 76.2 & 76.4 & 71.2 & 39.8 & 42.2 & 65.9 & 61.8 & 45.4 & 100.0 \\
        \midrule
        \rowcolor{c3light} \model 30B-A3B
        & 87.4
        & 82.7
        & 78.2
        & 60.2
        & 82.6
        & 91.0
        & 88.0
        & 88.0
        & 99.6 \\
        \rowcolor{c3light} \model 2B
        & 70.1 
        & 66.9 
        & 60.5 
        & 25.7 
        & 62.1 
        & 73.0 
        & 64.4 
        & 73.8 
        & 99.8  \\ 
    \bottomrule
    \end{tabular}
    \end{adjustbox}
    \end{TableSmall}
    \label{tab: voicebench}
\end{table}

%% file: sections/3_Related_Work.tex
\section{Related Work}
\label{sec:related_work}

\subsection{Unified Understanding and Generation Models}

It remains a challenging task to combine understanding and generation into a single LLM. This is caused by the nature of significant distinctions between these tasks: generation requires synthesizing fine-grained details, while understanding is generally achievable with succinct semantic features. Furthermore, it remains a nascent, open question on how to preserve strong text reasoning quality while adding unified multimodal capabilities. 

A number of studies have been conducted in the vision domain \citep{ye2026understanding,zhang2025unified,wu2026liquid,team2024chameleon,tang2025ugen,chen2025janus,wu2025vila,wang2024emu3,zhuang2025vargpt}. In summary, while it is possible to integrate all the text and multimodal capabilities into a single model, it is challenging to optimize all of them simultaneously -- gains on certain tasks may lead to degradations on other tasks (especially the text reasoning quality). Several works demonstrate that fusing multimodal data in early stages (such as pre-training) helps unified modeling \citep{team2024chameleon,team2026kimi}, and the MoE architecture also helps \citep{tong2026beyond}. Several works use diffusion heads to handle the generation part \citep{xie2025show,zhou2025transfusion,deng2025emerging,chen2025blip3}; this approach is orthogonal to our work.

The unification of understanding and generation is less studied in the general audio domain. Our previous work, UALM \citep{tian2025ualm}, took a first step towards this direction. With proper data blending and training recipes, UALM-7B achieves the state-of-the-art text-to-audio generation, matches the strong audio understanding models at the same size, and has small degradation on several text benchmarks including MMLU \citep{hendrycks2020measuring}, GSM8K \citep{cobbe2021training}, and HumanEval \citep{chen2021evaluating} compared to the base LLM. However, UALM's study on text benchmarks have limitations in terms of the following two aspects: $(i)$ the LLM evaluation domain is rapidly evolving and these three benchmarks get saturated, and $(ii)$ UALM only studies the effect of text SFT without text RL, and as a consequence, may exhibit limited performance on complex multi-step reasoning tasks.

In this work, we conduct a systematic study along this line and present \model. Compared to UALM, \model further improves text-to-audio and audio understanding quality, supports text-to-speech generation and speech recognition, and achieves systematically better text reasoning abilities along with a comprehensive study on the effect of text SFT+RL. Notably, \model is comparable with the base LLM (Nemotron-Cascade 2 \citep{Nemotron_Cascade_2}) on a wide range of text reasoning tasks including knowledge, chat, long context, and agentic benchmarks. We present comprehensive and effective training recipes for audio and text modeling that lead to our results.

\subsection{Speech LLMs}

Speech LLMs are a class of audio LLMs specialized for speech applications, including automatic speech recognition (ASR), automatic speech translation (AST), speech understanding, text-to-speech (TTS) generation, speech-to-speech (S2S) generation, and combinations of these tasks \citep{zhang2023speechgpt,radford2022whisper,wang2023neural,zhang2023speak,chen2024vall,wang2023viola,rubenstein2023audiopalm,wang2024speechx,huang2023makeavoice,kharitonov2023speak,nguyen2025spirit,kim2024clam,wang2025spark,defossez2024moshi}. 

The speech outputs of speech LLMs are usually discrete speech tokens obtained from codec models \citep{zeghidour2021soundstream,defossez2022high,kumar2023high,yang2023hifi,zhang2024speechtokenizer,defossez2024moshi,ji2025wavtokenizer,ye2025codec,ye2025llasa,gong2026moss}. Codec models are trained to encode speech waveforms into discrete tokens (or codebooks) and reconstruct waveform from these tokens, and are usually trained with the VQ-VAE style \citep{van2017neural} Encoder-Quantizer-Decoder architectures. For efficient representation, higher quality, and better latency, speech LLMs adopt various quantization methods such as residual vector quantization (RVQ) \citep{zeghidour2021soundstream} and finite scalar quantization (FSQ) \citep{mentzer2023finite}, and different codebook recipes and parallelized token patterns associated with them \citep{copet2024simple}. Some works also use a separate talker for better scalability and control \citep{xu2025qwen2,xu2025qwen3omni}.

The inputs to speech LLMs are usually in various formats, including discrete tokens from speech codec models \citep{wang2023neural, chen2024vall, zhang2025mimo}, continuous representations of an audio encoder \citep{liu2025voxtral, goel2025audio, lu2026desta2}, or a hybrid format \citep{ding2025kimi}. These approaches usually have different trade-offs between improving the recognition/understanding quality versus reducing input-output space mismatch.

There is another line of research called duplex models \citep{leviathan2018google,defossez2024moshi, hu2025salm, casanova2025open, chien2026moshirag, roy2026personaplex,bytedanceseed2026seeduplex}. These models focus on real-time interaction between the user and the agent, where both can speak-while-listen at the same time. Duplex models are based on speech LLMs and introduce additional channels for interaction, which is outside the scope of this paper.

\subsection{Audio Understanding Models}
Audio understanding models extend speech LLMs to the domain of general non-speech sound and music, and focus on semantic understanding of these sound. The tasks include captioning, question-answering, and classification in a wide range of perspectives. Especially, some works focus on non-speech sound understanding \citep{deshmukh2023pengi,ghosh2024gama,kong2024audio,ghosh2025audio,deshmukh2026mellow}, music understanding \citep{won2024foundation,ma2024foundation,deng2024musilingo,gardner2023llark,liu2024music,ghosh2025music}, or all types of sound including speech \citep{ghosh2026audio,goel2025audio,tangsalmonn,gong2023listen,xu2025qwen2,xu2025qwen3omni,qwen3.5_omni,ding2025kimi,zhang2025mimo,liu2025voxtral,wu2025step,tian2025step}. Some works also have speech outputs as discussed in the previous section \citep{xu2025qwen2,xu2025qwen3omni,qwen3.5_omni,wu2025step,tian2025step,ding2025kimi,zhang2025mimo}.

Most of the audio understanding models take continuous representations from one or more audio encoders as inputs. There are many types of audio encoders used based on the focus and tasks of the audio understanding model \citep{radford2022whisper,goel2025audio,chu2024qwen2audio,xu2025qwen3omni,nvidia2025canary1bv2,chen2022hts,chen2022beats,gong2021ast,CLAP2023,wu2023large,li2024mert,dhariwal2020jukebox}. Besides continuous representations, several works investigated audio understanding with discrete audio codecs, but there is still a gap from the state-of-the-art models in terms of audio understanding quality \citep{yang2024uniaudio, zhang2025mimo}.

\subsection{Audio Generation Models}

Unlike speech generation models where the goal is to speak the transcriptions, audio generation models (or text-to-audio models) focus on generating an audio clip that is semantically consistent with a given text prompt (e.g., \textit{The sound of river flows and birds chirping}). This task is similar to text-to-image generation \citep{ramesh2021zero,rombach2022high} applied to the general audio domain. 

Some early approaches model text-to-audio generation with an LLM to predict discrete audio tokens \citep{borsos2023audiolm, copet2024simple, kreuk2022audiogen, agostinelli2023musiclm, liu2025songgen}. However, recent leading approach is largely dominated by diffusion-based models \citep{liu2023audioldm,chen2024musicldm,huang2023make, evansfast, evans2024stable, lee2024etta, ghosal2023text,majumder2024tango,hung2024tangoflux} in terms of sample quality, controllability, duration, and inference speed. 

Our previous work, UALM \citep{tian2025ualm}, aims to close the long-standing quality gap using LLM-based methods for audio generation. With proper data scaling, classifier-free guidance \citep{ho2022classifier} at training and inference time, and a post-training curriculum, UALM presents state-of-the-art text-to-audio results using 1.5B and 7B LLMs. 
In this work, we leverage the techniques developed in UALM, and further scale to our 30B MoE LLM backbone \citep{nemotron_nano_v3}.

%% file: sections/Appendix_Benchmarks.tex
\section{Benchmarks, Evaluation Setups, and Baselines}
\label{appendix:benchmarks}

We include the details of benchmarks, evaluation setups, and baselines to make this technical report self-contained.

In this work, we use the same evaluation setups as Nemotron-Cascade 2~\citep{Nemotron_Cascade_2} for text benchmarks (see Appendix \ref{app: benchmark reasoning} -- \ref{app: benchmark agent}). The evaluation setups for audio tasks are in Appendix \ref{app: benchmark TTS} -- \ref{app: benchmark TTA}; in these tasks, the audio understanding evaluations follow Audio Flamingo 3 \citep{goel2025audio}, and the text-to-audio evaluations follow ETTA \citep{lee2024etta}.

\subsection{Reasoning}
\label{app: benchmark reasoning}

For math and code reasoning, we include 
\begin{itemize}[leftmargin=2em]
    \item
    \textbf{AIME 2025 \& 2026}~\citep{aime2025, aime2026}: 30 \& 30 problems from American Invitational Mathematics Examination at 2025 and 2026, respectively.
    \item
    \textbf{HMMT Feb 2025}~\citep{hmmt_feb_2025}: 30 problems from Harvard-MIT Mathematics Tournament 2025 February math competition.
    \item
    \textbf{IMO-AnswerBench}~\citep{luong2025towards}: 400 problems with verifiable answers carefully chosen from past Olympiad competitions
    and then altered by experts to avoid memorization.
    \item
    \textbf{LiveCodeBench}~\citep{jain2024livecodebench} contains diverse algorithm coding problems with unit tests, collected from AtCoder, LeetCode platforms. We evaluate models competitive coding capability on LiveCodeBench v6~(2024/08-2025/05, \textbf{454} problems in total). We report pass@1 accuracy in \emph{thinking} mode, averaged over 8 generations (avg@8).
    \item 
    \textbf{SciCode}~\citep{tian2024scicode} serves as a challenging benchmark to evaluate model's ability on solving realistic scientific research tasks from STEM domains. It contains {338} subproblems from {80} main tasks.
\end{itemize}

For \model-30B-A3B and Nemotron-Cascade-2-30B-A3B evaluated on AIME 2025, AIME 2026, HMMT 2025 Feb, and LiveCodeBench v6, we set the thinking budget (maximum response length) to 131K tokens, the sampling temperature to 1.0, the top-p value to 1.0. 
For the with-tool setting, we enable tool use by appending a system-prompt postfix, allowing the model to call a stateful Python executor for up to 100 tool calls with a maximum response length of 131K tokens.
For IMO-AnswerBench, we set to 256K tokens because we found the questions are significantly more difficult. We use and report the LLM-Judge score using GPT-OSS-120B~\citep{agarwal2025gpt} as the judge and the AnswerAutoGrader prompt~\citep{luong2025towards} for answer correctness on IMO-AnswerBench as the short answers are complicated for rule-based verifier to compute. Following \citet{liu2024acemath,liu2025acereason}, we report avg@64 for AIME/HMMT and avg@16 for IMO-AnswerBench.

For baseline models, we use official numbers from their reports if the official numbers are available.
Otherwise, we evaluate baseline models with the recommended settings, ensuring a thinking budget of at least 128K tokens.

\subsection{Knowledge}
For knowledge reasoning, we include:
\begin{itemize}[leftmargin=2em]
    \item 
    \textbf{MMLU-Redux}~\citep{gema2024mmlu} is a benchmark consisting of a subset of 3,000 manually re-annotated questions across 30 MMLU subjects~\citep{hendrycks2020measuring}, which eliminates the original annotation errors.
    We evaluate the models in \emph{thinking} mode and, due to the large test set size, report exact match~(EM) accuracy based on a single generation per question.
    \item 
    \textbf{MMLU-Pro}~\citep{wang2024mmlu} is an enhanced version of the original MMLU benchmark that mitigates model saturation by expanding to over 12,000 graduate-level questions and increasing answer choices from four to ten. We report EM accuracy in \emph{thinking} mode using one generation per question.
    \item 
    \textbf{GPQA-Diamond}~\citep{rein2024gpqa} is a benchmark for assessing an LLM’s scientific reasoning capability. It consists of the highest quality 198 GPQA questions covering graduate-level physics, biology, and chemistry. We report pass@1 accuracy in \emph{thinking} mode, averaged over 8 generations per question (avg@8) to reduce variance.
\end{itemize}
For \model-30B-A3B and Nemotron-Cascade-2-30B-A3B evaluated on MMLU-Redux, MMLU-Pro and GPQA-Diamond in \emph{thinking} mode, we use a temperature of 1.0, a top-p value of 0.95, and a 128K-token thinking budget (maximum response length).

\subsection{Alignment}

For alignment tasks, we include:
\begin{itemize}[leftmargin=2em]
    \item 
    \textbf{ArenaHard 2.0}~\citep{li2024crowdsourced} is a human-preference alignment benchmark featuring 750 diverse and rigorous real-user prompts. 
    The dataset is specifically structured with 500 prompts targeting open-ended software engineering problems and complex mathematical questions, while the remaining 250 focus on creative writing.
    It uses an automatic LLM-as-Judge approach to estimate human preferences relative to a baseline model, enabling fully automated, low-cost, and fast evaluation without human intervention.  
    In our experiments, we report results without style control to allow for straightforward comparison with the officially reported numbers of other models. We evaluate the models in \emph{thinking} mode, and use GPT-4.1 as the automated judge. 
    \item \textbf{IFBench}~\citep{pyatkin2026generalizing} extends IFEval~\citep{zhou2023instruction} by introducing 58 new, diverse, and challenging verifiable out-of-domain instruction constraints. It provides a separate constraint list to ensure no overlap between training and test constraints, enabling evaluation of an LLM’s generalization ability. The test set contains 294 prompts. We report pass@1 accuracy in \emph{thinking} mode, averaged over 8 generations (avg@8).
\end{itemize}
For \model-30B-A3B and Nemotron-Cascade-2 evaluated on IFBench and ArenaHard in \emph{thinking} mode, we use a temperature of 0.6, a top-p value of 0.95, and a maximum response length of 32K tokens. 
For baseline models, we use officially reported results whenever available; if such results are absent, we evaluate them using their recommended inference configuration or the same settings as ours.

\subsection{Long Context}

For long context tasks, we include:
\begin{itemize}[leftmargin=2em]
    \item
    \textbf{AA-LCR}~\citep{artificialanalysis2025lcr} consists of 100 challenging text-based questions that require reasoning over multiple long, real-world documents, including company reports, government consultations, legal documents, and academic papers. Each sample contains a document set averaging approximately 100k tokens. The questions are designed such that answers cannot be directly retrieved from the documents and instead require reasoning across multiple sources of information. We report pass@1 accuracy in thinking mode, averaged over 16 generations (avg@16).
    
    \item 
    \textbf{LongBench v2}~\citep{bai2025longbench} contains 503 challenging multiple-choice questions with context lengths ranging from 8k to 2M words. The benchmark spans six task categories: single-document QA, multi-document QA, long in-context learning, long dialogue history understanding, code repository understanding, and long structured data understanding. The questions are designed to be difficult; even human experts equipped with document search tools may require substantial time to answer them correctly. We evaluate models in thinking mode and report pass@1 accuracy averaged over four generations (avg@4).
    
    \item 
    \textbf{NIAH@256K | 1M (Ruler Subset)} refers to the needle-in-a-haystack (NIAH) tasks from the RULER benchmark~\citep{hsieh2024ruler} at 256K and 1M context length. The NIAH test~\citep{niah_2023} assesses an LLM's long-context ability to retrieve a specific piece of information (the ``needle'') embedded within long distractor text (the ``haystack''). 
    The RULER benchmark defines four variants of this task: Single NIAH, Multi-keys NIAH, Multi-values NIAH, and Multi-queries NIAH. Following~\citet{blakeman2025nemotron}, we evaluate 100 instances from each category using a 1M-token context setting.
    Models are evaluated in reasoning-off mode, and we report pass@1 accuracy from a single generation (avg@1).
\end{itemize}

\subsection{Agentic Tasks}
\label{app: benchmark agent}
For agentic tasks, we include:
\begin{itemize}[leftmargin=2em]
    \item 
    \textbf{$\tau^2$-Bench}~\citep{barres2025tau2} evaluates multi-turn customer-service agents in environments with explicit policies, tool use, and shared world-state updates. We evaluate on the three official subsets: airline (50 examples), retail (114 examples), and telecom (114 examples). To keep the standard error within 1.5, we report avg@16 on airline and avg@8 on both retail and telecom.
    \item
    \textbf{SWE-bench Verified}~\citep{openai2024swe_verified} is a subset of the original test set from SWE-bench~\citep{jimenez2023swe}, consisting of 500 samples verified to be non-problematic by human annotators. We evaluate models in \emph{non-thinking} mode and report pass@1 accuracy, averaged over 4 generations per prompt~(avg@4).
    \item 
    \textbf{Terminal Bench 2.0}\citep{merrill2026terminal} is adopted for evaluating agents in terminal-based environments, which comprises of 89 human-validated tasks across specialized fields such as scientific computing, machine learning, and system administration. Moving beyond simple code generation, this benchmark focuses on end-to-end workflows, requiring agents to demonstrate proficiency in holistic operations like model training, system configuration, and software debugging rather than just producing isolated functions. We evaluate the model using the default Terminus-2 scaffolding. We report avg@5 task success rate.
\end{itemize}

For $\tau^2$-Bench evaluation, we adopt a \emph{latest-turn thought retention} policy for managing reasoning traces in multi-turn interactions: we retain the model's reasoning content after the most recent user turn, while discarding reasoning content from earlier turns. The official $\tau^2$-Bench evaluation code follows a \emph{no thought carry-over} policy, which removes all prior reasoning content; in our experiments, this evaluation setup consistently reduces scores by 3--5 points relative to latest-turn thought retention. We attribute this gap to train--test mismatch, since our SFT data for $\tau^2$-style interactions is constructed with the same latest-turn thought retention policy, which is also the thought-state management strategy used in Nemotron-3-Nano-v3 and DeepSeek-V3.2. For the telecom subset, we additionally modify the system prompt to emphasize the dual-control setting by repeating the instruction ``Make sure you guide the user through the steps, do not perform user-side actions yourself.'' three times. We also tested a \emph{full thought retention} policy, which preserves reasoning content from all previous turns and more closely matches RL training, but found it gives similar accuracy to latest-turn thought retention while incurring substantially longer contexts. We therefore report our final $\tau^2$-Bench results using latest-turn thought retention.

For SWE-bench Verified, we use the OpenHands scaffold~\citep{wang2025openhands} as the agentic coding evaluation framework. We adopt a full interaction retention policy for agent trajectories, preserving the complete history of tool calls, observations, and model outputs across turns. This includes prior file views, search results, executed commands, and intermediate patches, enabling the model to maintain state and reason effectively over long-horizon debugging processes. We set the maximum context length to 256K tokens and allow up to 200 turns, consistent with our execution-based agentic SWE-RL training configuration. Notably, this evaluation setup closely mirrors our training environment, as both rely on execution-based feedback and multi-turn interaction within the same tool-augmented scaffold. This alignment reduces train–test mismatch and enables the model to more effectively transfer learned behaviors, such as iterative debugging, hypothesis refinement, and tool-driven reasoning, to the evaluation setting.

\subsection{Text-to-Speech}
\label{app: benchmark TTS}

For TTS evaluation, we use the Seed-TTS-Eval English subset \citep{anastassiou2024seed}. It contains 1000 samples from Common Voice \citep{ardila2020common}. The template is: \textit{<|text to speech|> Generate speech for this transcription. \{TRANSCRIPTION\}}.

The baselines include:
\begin{itemize}
    \item State-of-the-art TTS models: Seed-TTS \citep{anastassiou2024seed}, Step-Audio-TTS-3B \footnote{https://huggingface.co/stepfun-ai/Step-Audio-TTS-3B}, E2 TTS \citep{eskimez2024e2}, F5 TTS \citep{chen2024f5}, and
    \item General-purpose audio language models: Qwen2.5 Omni \citep{xu2025qwen2}, Qwen3 Omni \citep{xu2025qwen3omni}, MiMo-Audio \citep{zhang2025mimo}. 
\end{itemize}

\subsection{Speech Recognition and Speech-to-Text Translation}

For ASR and AST tasks, we include:

\begin{itemize}[leftmargin=2em]
    \item
    \textbf{Open ASR}~\citep{srivastav2025openasr} evaluates English ASR on AMI, Earnings22, GigaSpeech, LibriSpeech, SPGISpeech, TED-LIUM, and VoxPopuli, spanning meetings, earnings calls, audiobooks, podcasts/web audio, financial speech, talks, and parliamentary speech. We report WER using the official English normalization pipeline.\footnote{\url{https://github.com/huggingface/open_asr_leaderboard/tree/main/normalizer}}

    \item
    \textbf{Noisy ASR} evaluates robustness to additive noise on LibriSpeech test-clean. Similar to the noisy ASR evaluation in Canary-1B-v2~\citep{nvidia2025canary1bv2}, we synthesize noisy speech by mixing each clean utterance with a randomly selected noise or background sound at SNR levels of $-5$, $0$, $5$, $10$, $25$, $50$, and $100$ dB. We report WER using the same English normalization pipeline as Open ASR.

    \item
    \textbf{Multilingual Open ASR}~\citep{srivastav2025openasr} evaluates multilingual speech recognition. We use the FLEURS subset, including the five FLEURS languages in the leaderboard setting (German, Spanish, French, Italian, and Portuguese), and additionally report Korean and Russian. We report WER using the multilingual Open ASR normalization pipeline.\footnotemark[\value{footnote}]

    \item
    \textbf{FLEURS speech-to-text translation}~\citep{conneau2023fleurs} evaluates X$\rightarrow$English speech translation. We use the same seven source languages as multilingual ASR and report BLEU \citep{papineni2002bleu} and COMET~\citep{rei2020comet}. \footnote{COMET is computed with \texttt{unbabel-comet} v2.2.7 and the \texttt{Unbabel/wmt22-comet-da} checkpoint. \url{https://github.com/Unbabel/COMET}}
\end{itemize}

For OpenASR and Noisy ASR evaluation, \model and all speech-LM baselines are evaluated with ASR prompts without the language id. The prompt for \model is: \textit{\textbf{<sound>}$\backslash$nTranscribe the speech in the input audio}. For multilingual ASR, we use language ids in the prompts for all models including \model. For \model, we add "\textit{Language: \{LANGUAGE\_ID\}}" to the previous prompt.

For AST, we specify the source language in user prompts for baseline models, but do not use the source language for \model. The prompt for \model is: \textit{\textbf{<sound>}$\backslash$nTranslate the spoken content in the audio to English}. 

Additional details on prompting, decoding, and baseline-specific inference recipes are provided in Appendix~\ref{app:asr_ast_details}.

The baselines include:
\begin{itemize}
    \item State-of-the-art ASR/AST models: Whisper-large-v3 \citep{radford2022whisper}, Seamless-m4t-v2-large \citep{barrault2023seamless}, Canary-1B v2 \citep{nvidia2025canary1bv2}, Parakeet-TDT-0.6B-v3 \citep{nvidia2025canary1bv2}, Canary-Qwen-2.5B \footnote{https://huggingface.co/nvidia/canary-qwen-2.5b}, and
    \item General-purpose audio language models: Step-Audio-R1.1 \citep{tian2025step}, Qwen3 Omni \citep{xu2025qwen3omni}, MiMo-Audio \citep{zhang2025mimo}, Kimi-Audio \citep{ding2025kimi}, Voxtral-Small \citep{liu2025voxtral}. 
\end{itemize}

\subsection{Speech Interaction}

We evaluate on VoiceBench \citep{chen2026voicebench} and BigBenchAudio \citep{srivastava2022beyond,suzgun2022challenging} for speech interaction tasks. We use a cascaded pipeline (ASR, text reasoning, and TTS) to evaluate on these tasks, but different from prior cascaded systems that use different models (e.g. Whisper + GPT4o + TTS-1), each step of our pipeline uses the same \model checkpoint with the recommended inference recipe as described earlier. In detail, we first run \model to get transcriptions similar to ASR evaluation. We then use \model to run text reasoning similar to Reasoning and Knowledge benchmarks, with the following prompt: \textit{\{TRANSCRIBED QUESTION\} $\backslash$n$\backslash$n Conclude your response with the sentence `The answer is $\backslash\backslash$boxed\{\{X\}\}.`} After text reasoning is done, we extract the answer from the \textit{$\backslash\backslash$boxed\{\{X\}\}} outputs with some regex rules. For BigBenchAudio, we then generate speech samples using the transcription: \textit{The answer is \{ANSWER\}}.

The baselines are taken from the BigBenchAudio Leaderboard \footnote{\url{https://artificialanalysis.ai/speech-to-speech\#speech-reasoning}} and VoiceBench Leaderboard \footnote{\url{https://matthewcym.github.io/VoiceBench}}, and we additionally evaluate Kimi-Audio-Instruct \citep{ding2025kimi} and MiMo-Audio \citep{zhang2025mimo} on these benchmarks.

\subsection{Audio Understanding}

For audio understanding evaluation, we evaluate on the following benchmarks:
\begin{itemize}
    \item MMAU (v05.15.25) \citep{sakshi2024mmau}, MMAR \citep{ma2025mmar}, and MMSU \citep{wang2025mmsu} for general sound, speech, and music understanding. These benchmarks are all multiple choice questions. The prompt we use to evaluate \model is: \textit{\textbf{<sound>}$\backslash$n\{QUESTION\} Choose the correct option from the following options:$\backslash$n(A) \{CHOICE A\}$\backslash$n(B) \{CHOICE B\}$\backslash$n(C) \{CHOICE C\}$\backslash$n(D) \{CHOICE D\}.}
    
    \item Audio Entailment \citep{audioentail} on Clotho-v2 and AudioCaps for deductive reasoning. These benchmarks contain closed-ended questions and require outputs to be within \{yes, no, maybe\}. The prompt we use to evaluate \model is: \textit{\textbf{<sound>}$\backslash$nIs the following description true about the input audio? Answer in Yes, No or Maybe.$\backslash$nDescription: \{DESCRIPTION\}.}
    
    \item CMM \citep{leng2026curse} for hallucination detection. This benchmark also contains closed-ended questions with a focus on sound event existence, and requires outputs to be within \{yes, no\}. The prompt we use to evaluate \model is: \textit{\textbf{<sound>}$\backslash$nDid you hear \{SOUND EVENT\} in the audio?}
\end{itemize}

The baselines include Audio Flamingo 3 \citep{goel2025audio}, Step-Audio-R1.1 \citep{tian2025step}, Kimi-Audio \citep{ding2025kimi}, Voxtral-small \citep{liu2025voxtral}, MiMo-Audio \citep{zhang2025mimo}, and Qwen3-Omni-30B-A3B instruct/thinking models \citep{xu2025qwen3omni}.

\subsection{Text-to-Audio}
\label{app: benchmark TTA}

For TTA evaluation, we use the AudioCaps \citep{kim2019audiocaps} and SongDescriber \citep{manco2023thesong} benchmarks. We evaluate on 10-second clips following the evaluation setup in ETTA \citep{lee2024etta}. The template is: \textit{<|text to audio|> Generate audio for this caption. \{CAPTION\}}.

We report the Fréchet Distance measured in OpenL3 feature space \citep{cramer2019look}, noted as following prior works \citep{evans2024stable,lee2024etta}. This metric, noted as $\mathrm{FD}_{\mathrm{openl3}}$, measures the distance between the generated audio distribution and the reference set distribution, and is therefore the lower the more statistically realistic.

The baselines include:
\begin{itemize}
    \item Diffusion models: AudioLDM2 Large \citep{liu2024audioldm}, Stable Audio Open \citep{evans2024stable}, ETTA \citep{lee2024etta}, Tango2 \citep{majumder2024tango}, and TangoFlux \citep{hung2024tangoflux};
    \item Autoregressive models tailored for audio generation: MusicGen \citep{copet2024simple}, Magenta Realtime \citep{team2025live} \footnote{MusicGen and Magenta Realtime are trained on licensed corpora.}, UALM-Gen \citep{tian2025ualm}, and
    \item Versatile LLMs with audio generation capabilities: UALM \citep{tian2025ualm}.
\end{itemize}

%% file: sections/Appendix.tex
\section{Detailed Results Across Training Stages}
\label{app: results by stages}

Table \ref{tab: multi-stage middle results} includes \model-30B-A3B results at intermediate training stages. 
The \textit{Audio Gen. SFT} stage achieves the best AudioCaps results, along with very good SongDescriber and Seed-TTS-Eval results.  After the \textit{Audio Gen. + Audio Und. SFT} stage, TTS and SongDescriber results are better, and the model has strong audio understanding, ASR, and AST abilities. In both stages, the text abilities are kept across a variety of benchmarks. 

During the RL stage, we observe that the text benchmarks improve in correspondence to the specific RL stages. Meanwhile, we do not see huge degradations in audio-related benchmarks. The only exception is that MOPD degrades ASR and MMSU, but the results get recovered after two more RL stages. The other audio benchmarks are mostly stable during these RL stages, verifying that Nemotron Cascade-2 RL on specific text domains does not affect multimodal abilities much. 

\begin{table}[!h]
    \setlength{\cmidrulewidth}{\heavyrulewidth}
    \centering
    \caption{\model-30B-A3B Intermediate Stage Results. LS represents LibriSpeech ASR and is evaluated with WER. Fleurs (xx$\rightarrow$en)$_4$ is the averaged BLEU|COMET scores of AST across four languages (de, fr, it, es). Seed-TTS-Eval is evaluated with WER, where the numbers in parenthesis indicate fixed-voice TTS. AudioCaps and SongDescriber are evaluated with OpenL3 Fréchet Distance. We mark tool-integrated reasoning results with $\diamondsuit$ for math and code reasoning.}
    \begin{TableSmall}
    \begin{adjustbox}{max width=\textwidth}
    \begin{tabular}{rccccc}
        \toprule
        \textbf{Training Stages} & \multicolumn{5}{c}{\textbf{Benchmarks Results at Different Stages}} \\ \cmidrule(lr){1-1} \cmidrule(lr){2-6}
        & \multicolumn{5}{c}{\multirow{2}{*}{\textbf{(Part 1) Text Benchmarks}}} \\ 
        & \\ \cmidrule(lr){2-6}
        & AIME 2025 & HMMT Feb25 & LiveCodeBench v6 & MMLU-Pro & GPQA-Diamond \\ \cmidrule(lr){2-2} \cmidrule(lr){3-3} \cmidrule(lr){4-4} \cmidrule(lr){5-5} \cmidrule(lr){6-6}
        Audio Gen. SFT   
        & 91.1|94.5$^\diamondsuit$ & 91.0|88.1$^\diamondsuit$ & 81.8|85.7 & 78.1 & 72.8 \\
        Audio Gen. + Audio Und. SFT  
        & 91.7|96.3$^\diamondsuit$ & 92.2|91.4$^\diamondsuit$ & 82.3|85.3 & 77.5 & 73.8 \\
        Instruction-Following RL 
        & --         & --         & --         & --   & --   \\
        Multi-domain RL 
        & --         & --         & --         & --   & --   \\
        MOPD 
        & 92.8|98.4$^\diamondsuit$ & 95.1|92.9$^\diamondsuit$ & 85.9|87.4 & 78.2 & 74.2 \\
        RLHF 
        & --         & --         & --         & --   & --   \\
        Long-context RL (\textbf{\model})
        & 91.2|98.3$^\diamondsuit$ & 92.2|93.8$^\diamondsuit$ & 85.3|86.2 & 78.9 & 74.9 \\

        \toprule
        & \multicolumn{5}{c}{\multirow{2}{*}{\textbf{(Part 2) Text Benchmarks}}} \\ & \\ \cmidrule(lr){2-6}
        
        & IFBench & ArenaHard v2 & AA-LCR  & $\tau^2$-Bench & MMLU-Redux \\ \cmidrule(lr){2-2} \cmidrule(lr){3-3} \cmidrule(lr){4-4} \cmidrule(lr){5-5} \cmidrule(lr){6-6}
        Audio Gen. SFT  
        & 50.9 & 67.7 & 23.7 & 51.1 & 85.5 \\
        Audio Gen. + Audio Und. SFT 
        & 52.4 & 59.4 & 27.0 & 52.4 & 86.3 \\
        Instruction-Following RL 
        & 78.0 & 52.2 & --   & --   & --   \\
        Multi-domain RL     
        & 78.5 & 48.8 & --   & 49.3 & --   \\
        MOPD             
        & 80.2 & 75.7 & 33.3 & 56.5 & 85.6 \\
        RLHF             
        & 77.7 & 81.7 & --   & 54.7 & --   \\
        Long-context RL (\textbf{\model}) & 77.8 & 81.6 & 39.3 & 57.2 & 86.4 \\
        
        \toprule
        & \multicolumn{5}{c}{\multirow{2}{*}{\textbf{(Part 3) Speech Benchmarks}}} \\ & \\ \cmidrule(lr){2-6}
        
        & LS (clean|others) $\downarrow$ & LS (noisy) $\downarrow$ & Fleurs (xx$\rightarrow$en)$_4$ & BigBenchAudio & Seed-TTS-Eval $\downarrow$ \\ \cmidrule(lr){2-2} \cmidrule(lr){3-3} \cmidrule(lr){4-4} \cmidrule(lr){5-5} \cmidrule(lr){6-6}
        Audio Gen. SFT              & N/S & N/S & N/S & N/S & 2.28 \\
        Audio Gen. + Audio Und. SFT             & 1.33|2.99 & 2.75 & 33.9|86.9 & 88.0 & 2.07(1.76) \\
        Instruction-Following RL    & 1.34|3.03 & 2.77 & 33.8|86.9 & 88.9 & (1.75)     \\
        Multi-domain RL             & 1.36|3.06 & 2.73 & 33.8|86.9 & 90.9 & (1.76)     \\ 
        MOPD                        & 1.32|3.19 & 3.32 & 33.8|86.9 & 90.0 & (1.80)     \\ 
        RLHF                        & 1.36|3.00 & 3.03 & 33.9|87.0 & 89.7 & (1.93)     \\
        Long-context RL (\textbf{\model}) & 1.34|3.06 & 2.92 & 33.8|87.0 & 90.0 & (1.70)     \\
        
        \toprule
        & \multicolumn{5}{c}{\textbf{\multirow{2}{*}{(Part 4) Audio Benchmarks}}} \\ & \\  \cmidrule(lr){2-6}
        
        & MMAU-mini & MMAR & MMSU & AudioCaps $\downarrow$ & SongDescriber $\downarrow$ \\ \cmidrule(lr){2-2} \cmidrule(lr){3-3} \cmidrule(lr){4-4} \cmidrule(lr){5-5} \cmidrule(lr){6-6}
        Audio Gen. SFT & N/S & N/S & N/S & 59.3 & 65.7 \\
        Audio Gen. + Audio Und. SFT & 76.3 & 62.7 & 63.3 & 60.9 & 58.1 \\
        Instruction-Following RL & 76.5	& 62.0 & 62.9 & 65.3 & 59.9 \\
        Multi-domain RL & 76.3 & 61.6 &	63.0 & 64.2 & 59.6 \\
        MOPD  & 77.2 & 63.2 & 60.9 & 64.2 & 61.3 \\
        RLHF  & 77.4 & 63.2 & 63.1 & 64.3 & 61.1 \\
        Long-context RL (\textbf{\model}) & 76.5 & 63.2 & 63.4 & 66.9 & 62.7 \\
    \bottomrule
    \end{tabular}
    \end{adjustbox}
    \end{TableSmall}
    \label{tab: multi-stage middle results}
\end{table}

\section{Ablation Studies}
\label{app: ablation}

\subsection{Audio Warmup}
\label{app: warmup}

In this section, we compare different audio warmup methods in multi-stage SFT. The first (used on \model) is to freeze text token embeddings and only update audio token embeddings. This is implemented by gradient masking of text token embeddings. Since LLM is also frozen, this approach does not affect the text part of the model. The second approach is to simply unfreeze both text and audio token embeddings and train on all tasks including text. We hypothesized this approach could jointly optimize all embeddings for better multi-modal training. The third approach is to combine the previous two: we apply the first and then the second. 

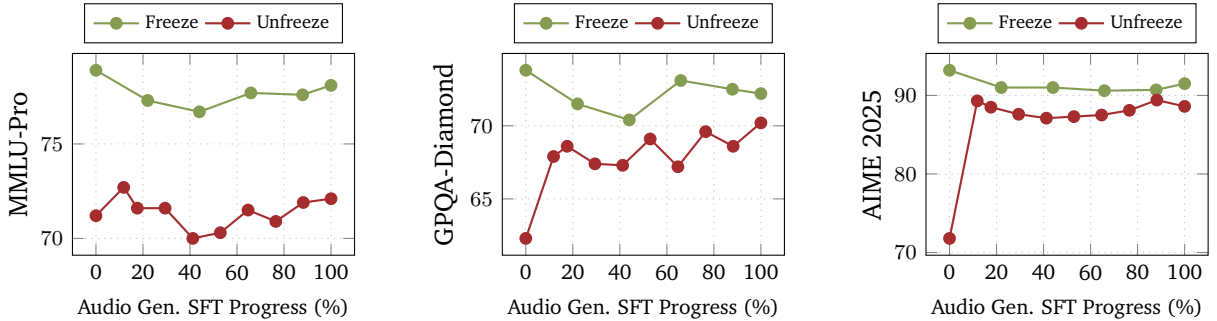
\begin{figure}[!h] 
    \centering
    \begin{subfigure}[t]{0.32\textwidth}
        \centering
        \captionsetup{width=0.9\linewidth}
        \begin{tikzpicture}
        \begin{axis}[
            width=\textwidth,
            height=0.8\textwidth,
            xlabel={\footnotesize Audio Gen. SFT Progress (\%)},
            xtick={0,20,40,60,80,100},
            xticklabels={0,20,40,60,80,100},
            ylabel={MMLU-Pro},
            ylabel style={yshift=-0.2em},
            grid=both,
            grid style={dotted, gray!50},
            mark options={solid},
            ticklabel style={font=\footnotesize},
            legend style={font=\scriptsize, at={(0.5,1.05)}, anchor=south, legend columns=2},
        ]
        \addplot[color=c3, thick, mark=*] coordinates {
            (0, 78.9)		(22, 77.3)		(44, 76.7)	(65.9, 77.7)	(87.9, 77.6)	(100, 78.1)
        };
        \addlegendentry{Freeze}
        
        \addplot[color=c5, thick, mark=*] coordinates {
            (0, 71.2)	(11.8, 72.7)	(17.6, 71.6)	(29.4, 71.6)	(41.2, 70)	(52.9, 70.3)	(64.7, 71.5)	(76.5, 70.9)	(88.2, 71.9)	(100, 72.1)
        };
        \addlegendentry{Unfreeze}
        \end{axis}
        \end{tikzpicture}
        
    \end{subfigure}
    \hfill
    \begin{subfigure}[t]{0.32\textwidth}
        \centering
        \captionsetup{width=0.9\linewidth}
        \begin{tikzpicture}
        \begin{axis}[
            width=\textwidth,
            height=0.8\textwidth,
            xlabel={\footnotesize Audio Gen. SFT Progress (\%)},
            xtick={0,20,40,60,80,100},
            xticklabels={0,20,40,60,80,100},
            ylabel={GPQA-Diamond},
            ylabel style={yshift=-0.2em},
            grid=both,
            grid style={dotted, gray!50},
            mark options={solid},
            ticklabel style={font=\footnotesize},
            legend style={font=\scriptsize, at={(0.5,1.05)}, anchor=south, legend columns=2},
        ]
        \addplot[color=c3, thick, mark=*] coordinates {
            (0, 73.8)		(22, 71.5)		(44, 70.4)	(65.9, 73.1)	(87.9, 72.5)	(100, 72.2)
        };
        \addlegendentry{Freeze}
        
        \addplot[color=c5, thick, mark=*] coordinates {
            (0, 62.3)	(11.8, 67.9)	(17.6, 68.6)	(29.4, 67.4)	(41.2, 67.3)	(52.9, 69.1)	(64.7, 67.2)	(76.5, 69.6)	(88.2, 68.6)	(100, 70.2)
        };
        \addlegendentry{Unfreeze}
        \end{axis}
        \end{tikzpicture}
        
    \end{subfigure}
    \hfill
    \begin{subfigure}[t]{0.32\textwidth}
        \centering
        \captionsetup{width=0.9\linewidth}
        \begin{tikzpicture}
        \begin{axis}[
            width=\textwidth,
            height=0.8\textwidth,
            xlabel={\footnotesize Audio Gen. SFT Progress (\%)},
            xtick={0,20,40,60,80,100},
            xticklabels={0,20,40,60,80,100},
            ylabel={AIME 2025},
            ylabel style={yshift=-0.2em},
            grid=both,
            grid style={dotted, gray!50},
            mark options={solid},
            ticklabel style={font=\footnotesize},
            legend style={font=\scriptsize, at={(0.5,1.05)}, anchor=south, legend columns=2},
        ]
        \addplot[color=c3, thick, mark=*] coordinates {
            (0, 93.2)		(22, 91)		(44, 91)	(65.9, 90.6)	(87.9, 90.7)	(100, 91.5)
        };
        \addlegendentry{Freeze}
        
        \addplot[color=c5, thick, mark=*] coordinates {
            (0, 71.8)	(11.8, 89.3)	(17.6, 88.5)	(29.4, 87.6)	(41.2, 87.1)	(52.9, 87.3)	(64.7, 87.5)	(76.5, 88.1)	(88.2, 89.4)	(100, 88.6)
        };
        \addlegendentry{Unfreeze}
        \end{axis}
        \end{tikzpicture}
        
    \end{subfigure}

    \caption{Audio Gen. SFT Results between Freezing and Unfreezing Text Embeddings in Audio Warmup. }
    \label{fig: freeze vs unfreeze text embeddings in audio warmup}
\end{figure}

The results after audio warmup are shown in Table \ref{tab: warmup comparison}. We find unfreezing text token embeddings has no benefit in any situation, and the gap to the first approach is huge. In Table \ref{fig: freeze vs unfreeze text embeddings in audio warmup}, we plot the text evaluation results of the Audio Gen. SFT stage after different warmup methods. If we freeze the text embeddings in Audio Warmup, then the model begins with high text scores and have small degradation at the beginning of Audio Gen. SFT in compensation for the audio generation tasks. If we unfreeze the text embeddings, then the model begins with lower text scores, recovers after 10\%-20\% of training, but is not comparable with the first approach.

\vspace{1em}
\begin{table}[!h]
    \centering
    \caption{Comparison of Audio Warmup Methods.}
    \begin{TableSmall}
    \begin{adjustbox}{max width=\textwidth}
    \begin{tabular}{l|ccc}
    \toprule
        Warmup Method & MMLU-Pro & GPQA-Diamond & AIME 2025 \\ \midrule
        Freeze text embeddings (\model's recipe) & 78.9 & 73.8 & 93.2 \\
        Unfreeze all embeddings & 71.2 & 62.3 & 71.8 \\
        Freeze text embeddings then unfreeze all embeddings & 65.1 & -- & -- \\ 
        Unfreeze all embeddings; higher text weights & 59.7 & -- & -- \\ 
    \bottomrule
    \end{tabular}
    \end{adjustbox}
    \end{TableSmall}
    \label{tab: warmup comparison}
\end{table}

\vspace{1em}

\subsection{Text Data Blending Ratios}
\label{app: text blending ratios}

In this section, we ablate the blending ratios for text data and investigate how this ratio affects text quality in the SFT training.

\begin{figure}[!h] 
    \centering
    \begin{subfigure}[t]{0.32\textwidth}
        \centering
        \captionsetup{width=0.9\linewidth}
        \begin{tikzpicture}
        \begin{axis}[
            width=\textwidth,
            height=0.8\textwidth,
            xlabel={\footnotesize Audio Gen. + Audio Und. SFT Progress (\%)},
            xtick={0,10,20,30,40},
            xticklabels={0,10,20,30,40},
            ylabel={MMLU-Pro},
            ylabel style={yshift=-0.2em},
            grid=both,
            grid style={dotted, gray!50},
            mark options={solid},
            ticklabel style={font=\footnotesize},
            legend style={font=\scriptsize, at={(0.5,1.05)}, anchor=south, legend columns=2},
        ]
        \addplot[color=c3, thick, mark=*] coordinates {
            (0, 78.1)	
            (14.7, 77.9)	
            (29.4, 77.6)
        };
        \addlegendentry{0.69}
        
        \addplot[color=c5, thick, mark=*] coordinates {
            (0, 78.1)	
            (16.7, 76.0)	
            (27.8, 76.2)
        };
        \addlegendentry{0.56}
        \end{axis}
        \end{tikzpicture}
        
    \end{subfigure}
    \hfill
    \begin{subfigure}[t]{0.32\textwidth}
        \centering
        \captionsetup{width=0.9\linewidth}
        \begin{tikzpicture}
        \begin{axis}[
            width=\textwidth,
            height=0.8\textwidth,
            xlabel={\footnotesize Audio Gen. + Audio Und. SFT Progress (\%)},
            xtick={0,10,20,30,40},
            xticklabels={0,10,20,30,40},
            ylabel={GPQA-Diamond},
            ylabel style={yshift=-0.2em},
            grid=both,
            grid style={dotted, gray!50},
            mark options={solid},
            ticklabel style={font=\footnotesize},
            legend style={font=\scriptsize, at={(0.5,1.05)}, anchor=south, legend columns=2},
        ]
        \addplot[color=c3, thick, mark=*] coordinates {
            (0, 72.2)	
            (14.7, 71.8)	
            (29.4, 72.3)
        };
        \addlegendentry{0.69}
        
        \addplot[color=c5, thick, mark=*] coordinates {
            (0, 72.2)	
            (16.7, 71.0)	
            (27.8, 70.3)
        };
        \addlegendentry{0.56}
        \end{axis}
        \end{tikzpicture}
        
    \end{subfigure}
    \hfill
    \begin{subfigure}[t]{0.32\textwidth}
        \centering
        \captionsetup{width=0.9\linewidth}
        \begin{tikzpicture}
        \begin{axis}[
            width=\textwidth,
            height=0.8\textwidth,
            xlabel={\footnotesize Audio Gen. + Audio Und. SFT Progress (\%)},
            xtick={0,10,20,30,40},
            xticklabels={0,10,20,30,40},
            ylabel={AIME 2025},
            ylabel style={yshift=-0.2em},
            grid=both,
            grid style={dotted, gray!50},
            mark options={solid},
            ticklabel style={font=\footnotesize},
            legend style={font=\scriptsize, at={(0.5,1.05)}, anchor=south, legend columns=2},
        ]
        \addplot[color=c3, thick, mark=*] coordinates {
            (0, 91.5)	
            (14.7, 91.3)	
            (29.4, 91.6)
        };
        \addlegendentry{0.69}
        
        \addplot[color=c5, thick, mark=*] coordinates {
            (0, 91.5)	
            (16.7, 90.1)	
            (27.8, 89.5)
        };
        \addlegendentry{0.56}
        \end{axis}
        \end{tikzpicture}
        
    \end{subfigure}
    \caption{Comparison of Text Data Blending Ratios (0.56 vs 0.69) in the Audio Gen. + Audio Und. SFT Stage.}
    \label{fig: multi-stage text ratio comparison}
\end{figure}
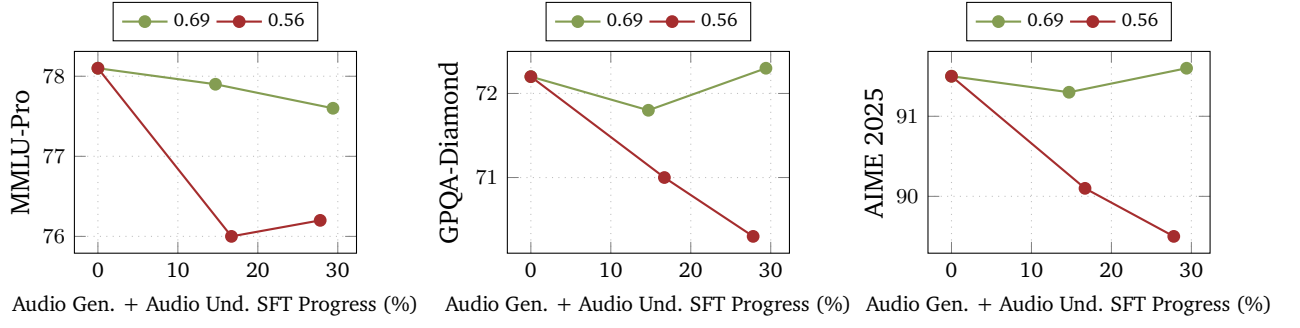

In Figure \ref{fig: multi-stage text ratio comparison}, we compare two text blending ratios: 0.56 vs 0.69 in the Audio Gen. + Audio Und. SFT stage. If the text blending ratio is 0.56 (close to 0.59 in the Audio Gen. SFT stage), there is considerable degradation on text evaluation. This is potentially because in the Audio Gen. + Audio Und. SFT stage, there are many audio-text pairs with text outputs; in contrast, in the Audio Gen. SFT stage, there is no text output for audio tasks, and therefore, we need a higher text blending ratio in the Audio Gen. + Audio Und. SFT stage to maintain the text quality. 

\begin{figure}[!h] 
    \centering
    \begin{subfigure}[t]{0.8\textwidth}
        \centering
        \captionsetup{width=0.9\linewidth}
        \begin{tikzpicture}
        \begin{axis}[
            width=\textwidth,
            height=0.3\textwidth,
            xlabel={\footnotesize Audio Gen. + Audio Und. + Text Consolidated SFT Progress (\%)},
            xtick={0,20,40,60,80,100},
            xticklabels={0,20,40,60,80,100},
            ylabel={MMLU-Pro},
            ylabel style={yshift=-0.2em},
            grid=both,
            grid style={dotted, gray!50},
            mark options={solid},
            ticklabel style={font=\footnotesize},
            legend style={font=\scriptsize, at={(0.5,1.05)}, anchor=south, legend columns=3},
        ]
        \addplot[color=c3, thick, dashed, mark=*] coordinates {
        (52.6, 78.2)
            (62.6, 77.5)	(67, 78.2)	(75.7, 78.8)	(84.5, 78.1)	(93.2, 78.6)		(100, 79.1)
        };
        \addlegendentry{0.88 (continued)}
        
        \addplot[color=c5, thick, dashed, mark=*] coordinates {
        (52.6, 78.2)
            (56.2, 76.2)	(58.4, 74.3)		(60.6, 73.7)		(64.2, 72.5)	(65.8, 62.7)
        };
        \addlegendentry{0.75 (continued)}

        \addplot[color=c0, thick, mark=*] coordinates {
            (26.3, 78.5)	(39.5, 78.4)	(49.4, 78.1)	(52.6, 78.2)
        };
        \addlegendentry{0.75 (initial)}
        \end{axis}
        \end{tikzpicture}
    \end{subfigure}
    \hfill
    \begin{subfigure}[t]{0.8\textwidth}
        \centering
        \captionsetup{width=0.9\linewidth}
        \begin{tikzpicture}
        \begin{axis}[
            width=\textwidth,
            height=0.3\textwidth,
            xlabel={\footnotesize Audio Gen. + Audio Und. + Text Consolidated SFT Progress (\%)},
            xtick={0,20,40,60,80,100},
            xticklabels={0,20,40,60,80,100},
            ylabel={GPQA-Diamond},
            ylabel style={yshift=-0.2em},
            grid=both,
            grid style={dotted, gray!50},
            mark options={solid},
            ticklabel style={font=\footnotesize},
            legend style={font=\scriptsize, at={(0.5,1.05)}, anchor=south, legend columns=3},
        ]
        \addplot[color=c3, thick, dashed, mark=*] coordinates {
        (52.6, 73)
            (62.6, 70.9)	(67, 70.6)	(75.7, 71.5)	(84.5, 71.2)	(93.2, 72.5)		(100, 72.8)
        };
        \addlegendentry{0.88 (continued)}
        
        \addplot[color=c5, thick, dashed, mark=*] coordinates {
        (52.6, 73)
            (56.2, 68.9)	(58.4, 67.7)		(60.6, 65.3)		(64.2, 65)	(65.8, 52.5)
        };
        \addlegendentry{0.75 (continued)}

        \addplot[color=c0, thick, mark=*] coordinates {
            (26.3, 72.2)	(39.5, 71.9)	(49.4, 73)		(52.6, 73)
        };
        \addlegendentry{0.75 (initial)}
        \end{axis}
        \end{tikzpicture}
    \end{subfigure}
    \hfill
    \begin{subfigure}[t]{0.8\textwidth}
        \centering
        \captionsetup{width=0.9\linewidth}
        \begin{tikzpicture}
        \begin{axis}[
            width=\textwidth,
            height=0.3\textwidth,
            xlabel={\footnotesize Audio Gen. + Audio Und. + Text Consolidated SFT Progress (\%)},
            xtick={0,20,40,60,80,100},
            xticklabels={0,20,40,60,80,100},
            ylabel={AIME 2025},
            ylabel style={yshift=-0.2em},
            grid=both,
            grid style={dotted, gray!50},
            mark options={solid},
            ticklabel style={font=\footnotesize},
            legend style={font=\scriptsize, at={(0.5,1.05)}, anchor=south, legend columns=3},
        ]
        \addplot[color=c3, thick, dashed, mark=*] coordinates {
        (52.6, 88.1)
            (62.6, 86.3)	(67, 88.9)	(75.7, 89.7)	(84.5, 89.9)	(93.2, 90.3)		(100, 90.6)
        };
        \addlegendentry{0.88 (continued)}
        
        \addplot[color=c5, thick, dashed, mark=*] coordinates {
        (52.6, 88.1)
            (56.2, 86.2)	(58.4, 85.6)		(60.6, 80.6)		(64.2, 78.7)	(65.8, 66.8)
        };
        \addlegendentry{0.75 (continued)}

        \addplot[color=c0, thick, mark=*] coordinates {
            (26.3, 89.4)	(39.5, 89.6)	(49.4, 88.5)	(52.6, 88.1)
        };
        \addlegendentry{0.75 (initial)}
        \end{axis}
        \end{tikzpicture}
    \end{subfigure}
    \caption{Comparison of Text Data Blending Ratios (0.75 vs 0.88) in the Single-Stage Consolidated SFT Stage.}
    \label{fig: single-stage text ratio comparison}
\end{figure}
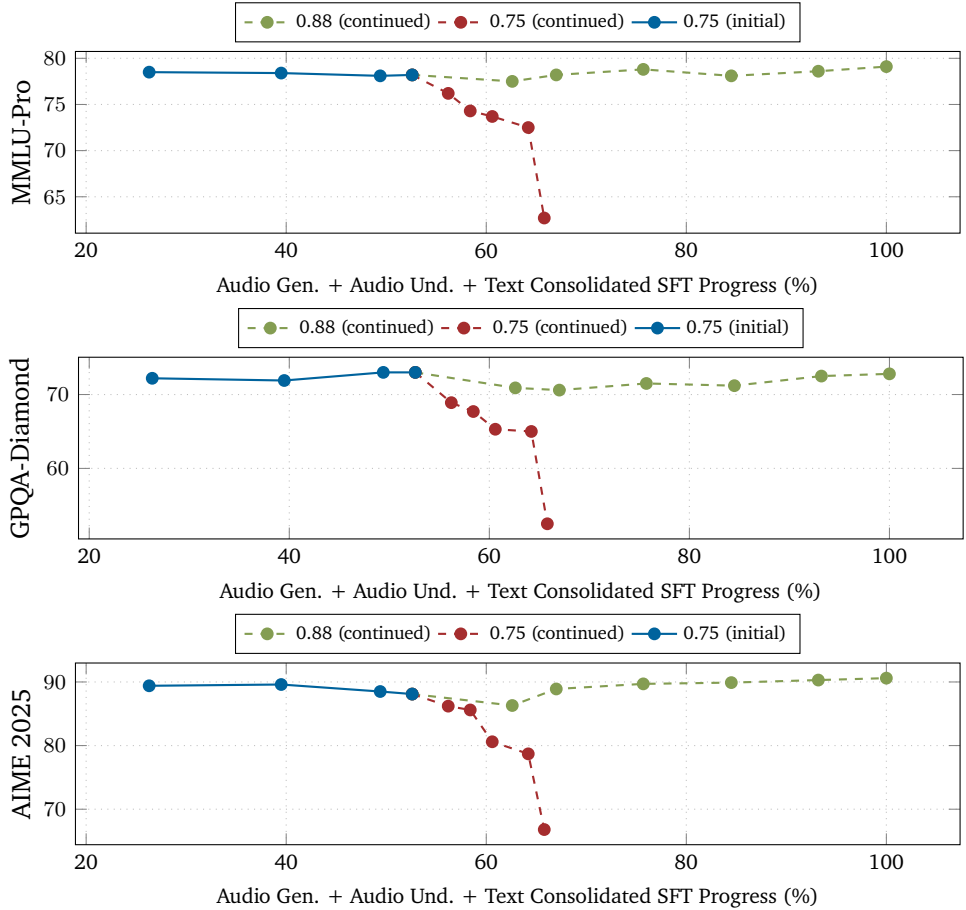

In Figure \ref{fig: single-stage text ratio comparison}, we demonstrate the text benchmark evaluations of the single-stage consolidated SFT model through the training progress. We begin with a text blending ratio of 0.75 (marked in \textcolor{c0}{blue}), but observe significant cliff of the text evaluation results around 50\%-60\% of training (marked in \textcolor{c5}{red}). We re-started from the checkpoint before the cliff (the last blue point) with different random seeds and observed cliff again around 60\%-80\% of training. We therefore increase the text blending ratio to 0.88 (marked in \textcolor{c3}{green}), re-start from the last blue point, and observe mild or no text degradations. We conclude that the single-stage consolidated SFT requires a significantly higher text blending ratio to keep the text quality, and this still does not alleviate the long-context issue discussed in Section \ref{sec: multi-stage vs single-stage SFT}.

\subsection{CFG Values and Hyper-parameters for TTS and TTA}
\label{app: cfg}

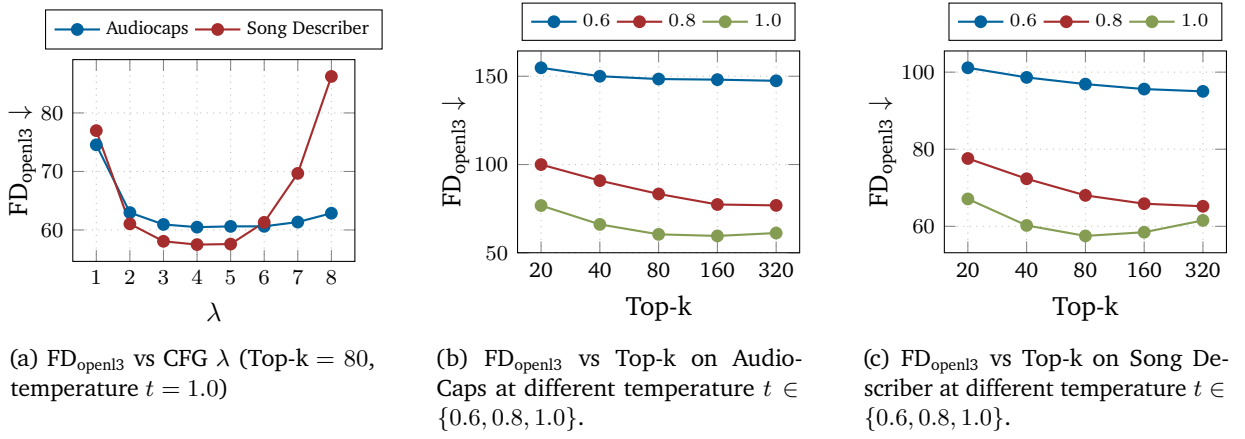
\begin{figure}[!h] 
    \centering
    \begin{subfigure}[t]{0.32\textwidth}
        \centering
        \captionsetup{width=0.9\linewidth}
        \begin{tikzpicture}
        \begin{axis}[
            width=\textwidth,
            height=0.8\textwidth,
            xlabel={$\lambda$},
            xtick={1,2,3,4,5,6,7,8},
            ylabel={FD$_\text{openl3}$ $\downarrow$},
            ylabel style={yshift=-0.5em},
            grid=both,
            grid style={dotted, gray!50},
            mark options={solid},
            ticklabel style={font=\footnotesize},
            legend style={font=\scriptsize, at={(0.5,1.05)}, anchor=south, legend columns=2},
        ]
        \addplot[color=c0, thick, mark=*] coordinates {
            (1, 74.55) (2, 62.97) (3, 60.94) (4, 60.48)
            (5, 60.61) (6, 60.62) (7, 61.35) (8, 62.86)
        };
        \addlegendentry{Audiocaps}
        
        \addplot[color=c5, thick, mark=*] coordinates {
            (1, 76.98) (2, 61.04) (3, 58.06) (4, 57.50)
            (5, 57.59) (6, 61.30) (7, 69.67) (8, 86.24)
        };
        \addlegendentry{Song Describer}
        \end{axis}
        \end{tikzpicture}
        \caption{FD$_\text{openl3}$ vs CFG $\lambda$ (Top-k $=80$, temperature $t=1.0$)}
    \end{subfigure}
    \hfill
    \begin{subfigure}[t]{0.32\textwidth}
        \centering
        \captionsetup{width=0.9\linewidth}
        \begin{tikzpicture}
        \begin{axis}[
            width=\textwidth, 
            height=0.8\textwidth,
            xlabel={Top-k},
            xmode=log,
            log basis x=2,
            xtick={20,40,80,160,320},
            xticklabels={20,40,80,160,320},
            ylabel={FD$_\text{openl3}$ $\downarrow$},
            ylabel style={yshift=-0.5em},
            grid=both,
            grid style={dotted, gray!50},
            mark options={solid},
            ticklabel style={font=\footnotesize},
            legend style={font=\scriptsize, at={(0.5,1.05)}, anchor=south, legend columns=3},
        ]
        \addplot[color=c0, thick, mark=*] coordinates {
            (20, 154.7849)
            (40, 149.9967)
            (80, 148.4170)
            (160, 148.0652)
            (320, 147.4055)
        };
        \addlegendentry{$0.6$}
        
        \addplot[color=c5, thick, mark=*] coordinates {
            (20, 99.9888)
            (40, 90.8887)
            (80, 83.3503)
            (160, 77.3837)
            (320, 76.8730)
        };
        \addlegendentry{$0.8$}
        
        \addplot[color=c3, thick, mark=*] coordinates {
            (20, 76.7972)
            (40, 66.1118)
            (80, 60.4825)
            (160, 59.5910)
            (320, 61.2184)
        };
        \addlegendentry{$1.0$}
        \end{axis}
        \end{tikzpicture}
        \caption{FD$_\text{openl3}$ vs Top-k on AudioCaps at different temperature $t\in\{0.6,0.8,1.0\}$.}
    \end{subfigure}
    \hfill
    \begin{subfigure}[t]{0.32\textwidth}
        \centering
        \captionsetup{width=0.9\linewidth}
        \begin{tikzpicture}
        \begin{axis}[
            width=\textwidth, 
            height=0.8\textwidth,
            xlabel={Top-k},
            xmode=log,
            log basis x=2,
            xtick={20,40,80,160,320},
            xticklabels={20,40,80,160,320},
            ylabel={FD$_\text{openl3}$ $\downarrow$},
            ylabel style={yshift=-0.5em},
            grid=both,
            grid style={dotted, gray!50},
            mark options={solid},
            ticklabel style={font=\footnotesize},
            legend style={font=\scriptsize, at={(0.5,1.05)}, anchor=south, legend columns=3},
        ]
        \addplot[color=c0, thick, mark=*] coordinates {
            (20, 101.1315)
            (40, 98.6368)
            (80, 96.8901)
            (160, 95.6082)
            (320, 95.0136)
        };
        \addlegendentry{$0.6$}
        
        \addplot[color=c5, thick, mark=*] coordinates {
            (20, 77.5972)
            (40, 72.3329)
            (80, 68.0208)
            (160, 65.8634)
            (320, 65.1957)
        };
        \addlegendentry{$0.8$}
        
        \addplot[color=c3, thick, mark=*] coordinates {
            (20, 67.0976)
            (40, 60.2277)
            (80, 57.5048)
            (160, 58.4648)
            (320, 61.5298)
        };
        \addlegendentry{$1.0$}
        \end{axis}
        \end{tikzpicture}
        \caption{FD$_\text{openl3}$ vs Top-k on Song Describer at different temperature $t\in\{0.6,0.8,1.0\}$.}
    \end{subfigure}

    \caption{Comparisons across different inference hyperparameters for text-to-audio generation. }
    \label{fig: inference hparam TTA}
\end{figure}

\begin{figure}[!h] 
    \centering
    \begin{subfigure}[t]{0.32\textwidth}
        \centering
        \captionsetup{width=0.9\linewidth}
        \begin{tikzpicture}
        \begin{axis}[
            width=\textwidth,
            height=0.8\textwidth,
            xlabel={$\lambda$},
            xtick={1,2,3,4,5,6},
            ylabel={WER $\downarrow$},
            ylabel style={yshift=-0.5em},
            grid=both,
            grid style={dotted, gray!50},
            mark options={solid},
            ticklabel style={font=\footnotesize},
            legend style={font=\scriptsize, at={(0.5,1.05)}, anchor=south, legend columns=2},
        ]
        \addplot[color=c0, thick, mark=*] coordinates {
            (1, 1.65) 
            (1.5, 1.5) 
            (2, 1.74)
            (3, 1.74)
            (4, 2.08)
            (5, 2.48)
            (6, 3.04)
        };
        \addlegendentry{Seed-TTS-Eval (en)}
        
        \end{axis}
        \end{tikzpicture}
        \caption{WER vs CFG $\lambda$ on Seed-TTS-Eval (Top-k $=80$, temperature $t=0.1$)}
    \end{subfigure}
    \hfill
    \begin{subfigure}[t]{0.32\textwidth}
        \centering
        \captionsetup{width=0.9\linewidth}
        \begin{tikzpicture}
        \begin{axis}[
            width=\textwidth, 
            height=0.8\textwidth,
            xlabel={Top-k},
            xmode=log,
            log basis x=2,
            xtick={20,40,80,160,320},
            xticklabels={20,40,80,160,320},
            ylabel={WER $\downarrow$},
            ylabel style={yshift=-0.5em},
            grid=both,
            grid style={dotted, gray!50},
            mark options={solid},
            ticklabel style={font=\footnotesize},
            legend style={font=\scriptsize, at={(0.5,1.05)}, anchor=south, legend columns=3},
        ]
        \addplot[color=c0, thick, mark=*] coordinates {
            (20, 1.70)
            (40, 1.61)
            (80, 1.51)
            (160, 1.60)
            (320, 1.63)
        };
        \addlegendentry{$0.05$}
        
        \addplot[color=c3, thick, mark=*] coordinates {
            (20, 1.58)
            (40, 1.56)
            (80, 1.50)
            (160, 1.57)
            (320, 1.55)
        };
        \addlegendentry{$0.10$}
        
        \addplot[color=c5, thick, mark=*] coordinates {
            (20, 1.63)
            (40, 1.64)
            (80, 1.59)
            (160, 1.74)
            (320, 1.63)
        };
        \addlegendentry{$0.15$}

        \addplot[color=c4, thick, mark=*] coordinates {
            (20, 1.54)
            (40, 1.60)
            (80, 1.63)
            (160, 1.60)
            (320, 1.57)
        };
        \addlegendentry{$0.20$}

        \addplot[color=c1, thick, mark=*] coordinates {
            (20, 1.62)
            (40, 1.65)
            (80, 1.65)
            (160, 1.60)
            (320, 1.69)
        };
        \addlegendentry{$0.30$}
        
        \end{axis}
        \end{tikzpicture}
        \caption{WER vs Top-k on Seed-TTS-Eval at different temperature $t\in\{0.05,0.10,0.15,0.20,0.30\}$.}
    \end{subfigure}
    \hfill
    \begin{subfigure}[t]{0.32\textwidth}
        \centering
        \captionsetup{width=0.9\linewidth}
        \begin{tikzpicture}
        \begin{axis}[
            width=\textwidth, 
            height=0.8\textwidth,
            xlabel={Temperature $t$},
            xtick={0.0,0.10,0.20,0.30},
            xticklabels={0.0,0.10,0.20,0.30},
            ylabel={WER $\downarrow$},
            ylabel style={yshift=-0.5em},
            grid=both,
            grid style={dotted, gray!50},
            mark options={solid},
            ticklabel style={font=\footnotesize},
            legend style={font=\scriptsize, at={(0.5,1.05)}, anchor=south, legend columns=3},
        ]
        \addplot[color=c0, thick, mark=*] coordinates {
            (0.05, 1.70)
            (0.10, 1.58)
            (0.15, 1.63)
            (0.20, 1.54)
            (0.30, 1.62)
        };
        \addlegendentry{20}
        
        \addplot[color=c1, thick, mark=*] coordinates {
            (0.05, 1.61)
            (0.10, 1.56)
            (0.15, 1.64)
            (0.20, 1.60)
            (0.30, 1.65)
        };
        \addlegendentry{40}
        
        \addplot[color=c3, thick, mark=*] coordinates {
            (0.05, 1.51)
            (0.10, 1.50)
            (0.15, 1.59)
            (0.20, 1.63)
            (0.30, 1.65)
        };
        \addlegendentry{80}
        
        \addplot[color=c4, thick, mark=*] coordinates {
            (0.05, 1.60)
            (0.10, 1.57)
            (0.15, 1.74)
            (0.20, 1.60)
            (0.30, 1.60)
        };
        \addlegendentry{160}
        
        \addplot[color=c5, thick, mark=*] coordinates {
            (0.05, 1.63)
            (0.10, 1.55)
            (0.15, 1.63)
            (0.20, 1.57)
            (0.30, 1.69)
        };
        \addlegendentry{320}
        
        
        
        \end{axis}
        \end{tikzpicture}
        \caption{WER vs Temperature $t$ on Seed-TTS-Eval at different Top-k $\in\{20,40,80,160,320\}$.}
    \end{subfigure}

    \caption{Comparisons across different inference hyperparameters for text-to-speech generation. }
    \label{fig: inference hparam TTS}
\end{figure}
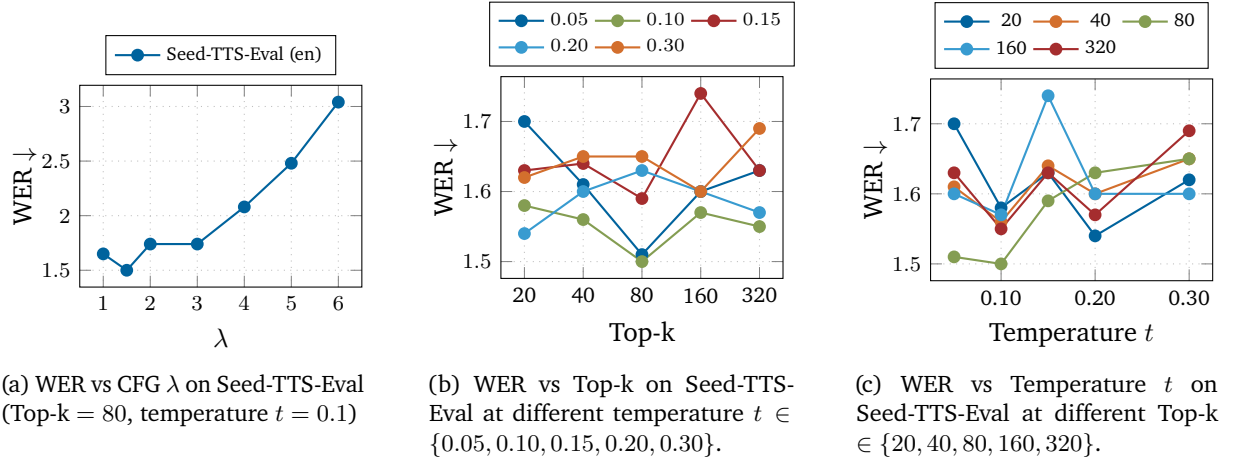

In this section, we study the optimal inference hyper-parameters for TTA (Figure \ref{fig: inference hparam TTA}) and TTS (Figure \ref{fig: inference hparam TTS}). We find for TTA, the best CFG value $\lambda\in[3,5]$ and use $\lambda=3$ in our experiments. We find the best Top-k to be 80 and temperature to be 1.0. For TTS, the best CFG value $\lambda\approx1.5$ and use $\lambda=1.5$ in our experiments. We also find the WER is close to optimal even without CFG. We find the best Top-k to be 80 and temperature to be 0.1.

~~\newpage
\section{Additional Details for ASR and AST Evaluations}
\label{app:asr_ast_details}

\model uses a chat-style template, where the user turn contains the task instruction and a \texttt{<sound>} placeholder, and a response in the assistant turn. For ASR, this response consists of language identification followed by the transcription. For AST, it consists of source-language identification, source transcription, and English translation. In language-conditioned multilingual ASR, we additionally teacher-force the language identification field to the known source language and let \model continue to generate the transcription. For AST, we \textit{do not} teacher-force the language field and instead let the model generate the full structured response.

For ASR-specific baselines, we use public results from the OpenASR leaderboard for Whisper Large v3, Canary-1B-v2, Parakeet-TDT-0.6B-v3, and Canary-Qwen-2.5B. For noisy and multilingual ASR evaluations of the baselines, we use the same model-specific ASR inference recipe that reproduces the OpenASR setting, with one small change for multilingual ASR that we provide the source-language information. Canary-1B-v2 and Parakeet-TDT-0.6B-v3 are evaluated on the European FLEURS languages they support, and Korean is excluded for these two models. For Canary-Qwen-2.5B, we only report English ASR results. Seamless-M4T v2 Large is reported for multilingual ASR and AST, using the source language as the generation target for ASR and English as the target for translation.

For speech/audio LLM baselines, we use task-specific prompts, summarized in Table~\ref{tab:speech_llm_prompt_setup}. For Qwen3-Omni-30B-A3B (Instruct / Thinking), we follow the official ASR and AST prompts. For the Thinking variant, we strip the reasoning trace and score only the final answer. Step-Audio R1.1 is a reasoning model, and we evaluate it in thinking mode with reasoning-budget control following the Nemotron 3 Nano Omni setup~\citep{deshmukh2026nemotron}. Specifically, we cap the thinking segment at 13000 tokens and allow a 1024-token grace budget after the reasoning trace, while using task-specific ASR and translation prompts. Kimi-Audio Instruct is evaluated on OpenASR with its official ASR prompt, but we omit multilingual ASR and AST because the released model and examples primarily target English and Chinese. Voxtral Small 24B is evaluated with its speech transcription and translation endpoints when applicable. For MiMo-Audio, we report the LibriSpeech test-clean number from the paper, as we did not identify a comparable full-benchmark ASR/AST evaluation path using the released inference setup.

\begin{table}[!h]
\centering
\small
\begin{tabular}{llp{0.58\linewidth}}
\toprule
Baseline & Setting & Prompting setup \\
\midrule
\multirow{3}{*}{Step-Audio R1.1 33B} &
Open / Noisy ASR &
``Transcribe the spoken audio.'' \\
&
Multilingual ASR &
``The audio language is <LANG>. Transcribe the spoken audio.'' \\
&
AST &
``The audio language is <LANG>. Translate the spoken audio into English.'' \\
\midrule
Kimi-Audio Instruct &
OpenASR &
``Please transcribe the following audio:'' \\
\midrule
\multirow{3}{*}{Qwen3-Omni-30B-A3B} &
Open / Noisy ASR &
``Transcribe the audio into text.'' \\
&
Multilingual ASR &
``Transcribe the <LANG> audio into text.'' \\
&
AST &
``Listen to the provided <LANG> speech and produce a translation in English text.'' \\
\midrule
\multirow{3}{*}{Voxtral Small 24B} &
Open / Noisy ASR &
``Transcribe the spoken content in the input speech. Only output the transcription.'' \\
&
Multilingual ASR &
``The audio language is <LANG>. Transcribe the spoken content in the input speech. Only output the transcription.'' \\
&
AST &
``The audio language is <LANG>. Translate the spoken content in the audio to English. Only output the English translation.'' \\
\bottomrule
\end{tabular}
\caption{Prompting setup used to reproduce speech/audio LLM baselines. <LANG> denotes the source language name.}
\label{tab:speech_llm_prompt_setup}
\end{table}